\documentclass{article}
\usepackage{arxiv}
\renewcommand{\undertitle}{}
\renewcommand{\headeright}{}
\renewcommand{\shorttitle}{}
\renewcommand{\headrulewidth}{0pt}
\usepackage{natbib}
\usepackage[T1]{fontenc}

\usepackage{hyperref}
\usepackage{url}
\usepackage{amsmath,amssymb,booktabs,graphicx}
\usepackage[table]{xcolor}
\usepackage[utf8]{inputenc}

\newcommand{\promptbox}[1]{%
  \par\vspace{0.45em}\noindent
  \colorbox[gray]{0.94}{\parbox{\dimexpr\linewidth-2\fboxsep\relax}{\ttfamily\small #1}}%
  \par\vspace{0.55em}}

\title{SelfCue: Making a 3D CT Report Generator\\ Say What It Already Knows}

\author{
\textbf{Renjie Liang\textsuperscript{1}, Yang Yang\textsuperscript{1}, Jinqian Pan\textsuperscript{1},} \\
\textbf{Zhengkang Fan\textsuperscript{1}, Chengkun Sun\textsuperscript{1}, Jie Xu\textsuperscript{1}} \\[4pt]
\small\textsuperscript{1}University of Florida, Gainesville, FL, USA
}
\date{}

\begin{document}

\maketitle


\begin{abstract}
Progress in 3D CT report generation is usually sought in increasingly sophisticated architectures and larger pools of training data. We find instead that a 3D CT report generator already holds what its report leaves out, and loses it when the hidden state becomes tokens. Over the 18 CT-RATE abnormalities, this hidden-to-report surfacing gap is reflected by a drop in macro AUROC from 0.848 in the hidden states to 0.739 in the generated report. We propose \textbf{SelfCue} based on contrastive decoding. It promotes what the hidden state already supports and suppresses what it does not. It raises clinical efficacy F1 to $0.481$ and the LLM-judged GREEN score to $0.510$. Distilling that behaviour into the weights gives \textbf{SelfCue-KD}, a student that keeps most of the gain, needs nothing extra at inference, and drops into any pipeline already serving the baseline. Code is available at \url{https://github.com/renjie-liang/SelfCue-CT}.
\end{abstract}

\section{Introduction}
\label{sec:intro}

Computed tomography is a workhorse of diagnostic imaging, with about $93$ million examinations performed in the United States in 2023 \citep{10.1001/jamainternmed.2025.0505}. Reporting one takes a radiologist about $15$ minutes \citep{kurmukov2024workload}. Automating that reading has produced a steady line of 3D CT models that write reports \citep{Hamamci_2026, liang2026adaptive}. They share one recipe. A 3D encoder turns the volume into visual tokens, a projector maps those tokens into a language model, and the model is fine-tuned to produce text. Their reports often get the abnormalities wrong, and fall short in overall quality. Most work has responded with stronger encoders \citep{wald2025colipri, Cao_2025_ICCV} or an explicit abnormality classifier attached alongside \citep{10981073, 10.1609/aaai.v38i3.28038}. We instead ask what the generator already encodes.

\begin{figure}[t]
\centering
\includegraphics[width=\linewidth]{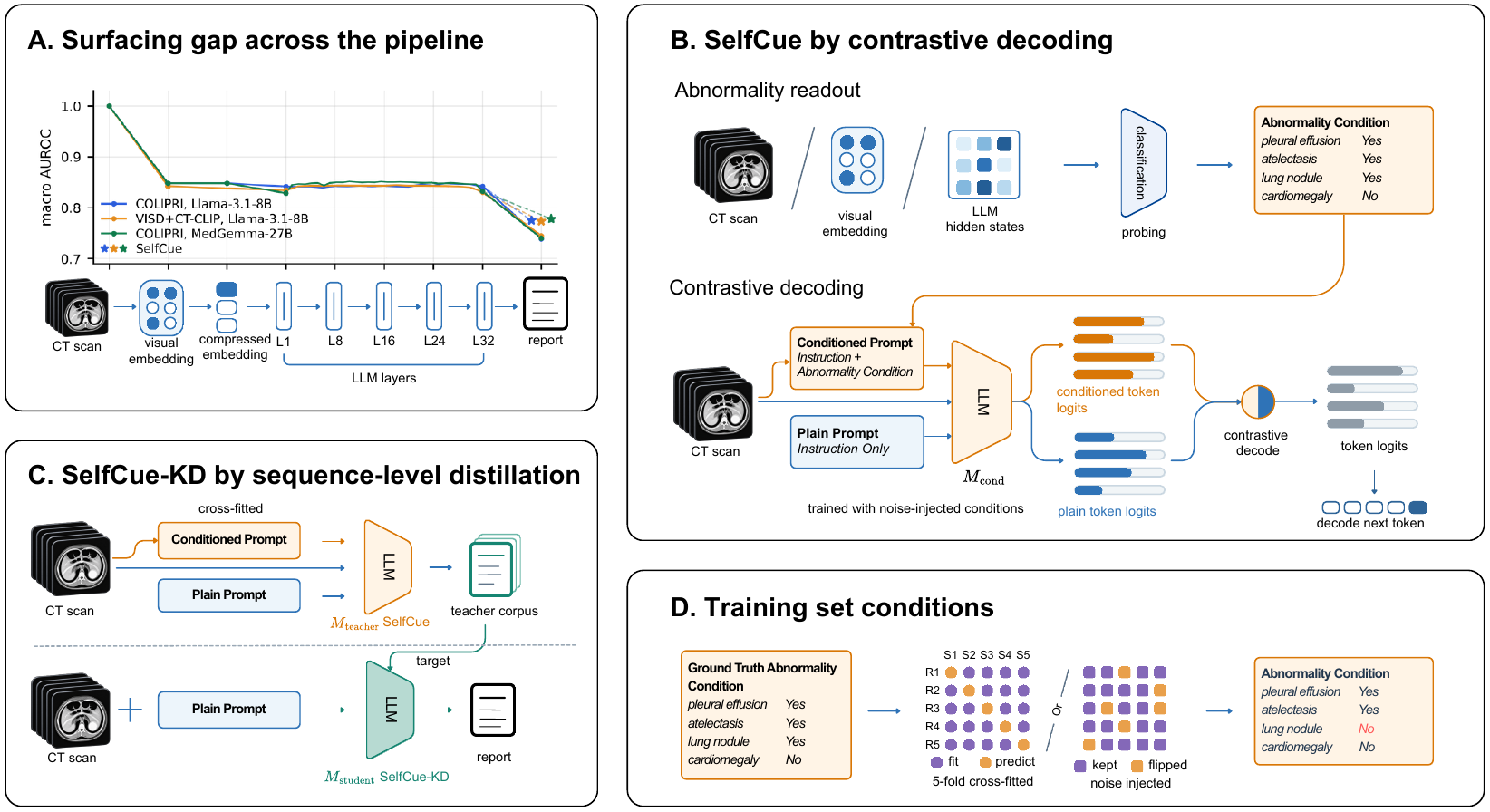}

\caption{\textbf{A.} Macro AUROC over the 18 CT-RATE abnormalities at each stage of three report generators, the raw volume $1.0$ by assumption. \textbf{B.} The abnormality condition is read by a probe, and each next token is chosen by contrasting the conditioned and plain branches. \textbf{C.} The teacher's reports train the student, which reads no condition. \textbf{D.} The two constructions of the training set conditions. Red marks a probe error.}
\label{fig:overview}
\end{figure}

We probe every stage of the pipeline on one ruler \citep{liang2026cheapprobespredictexpensive}, traced in Figure~\ref{fig:overview}A. For AUROC over the 18 CT-RATE abnormalities, the raw CT volume is $1.0$ by assumption, since the abnormality is physically present to be seen. It remains $0.848$ before the language model, and stays there at every layer we probe. It drops to $0.739$ in the generated report. A large part of the information therefore never reaches the text. The gap does not open gradually across the network but only when the hidden state becomes tokens, so what the report leaves out is missing from its text and not from the model that writes it. The same shape appears on two CT visual encoders and on two language models, so it is not a quirk of one pipeline. Our goal is to narrow this surfacing gap.

We propose \textbf{SelfCue} to correct the output distribution. We want the abnormalities the model already encodes to appear in the report, and the ones it does not to stay out. Following contrastive decoding \citep{pmlr-v235-sanchez24a,shi-etal-2024-trusting}, SelfCue runs the model twice for every token, once under a prompt carrying an abnormality condition and once without it. It decodes from a combination of the two that pushes away from the second. The abnormality condition is predicted for each volume, not read off the labels.

SelfCue has two costs at inference. It runs the model twice for every token, which doubles the work and does not fit standard inference engines. The abnormality condition also has to be computed for every volume before decoding. We distil the corrected decoder into a student model we call \textbf{SelfCue-KD}. Following sequence-level distillation \citep{kim-rush-2016-sequence}, SelfCue writes a report for every training volume, and the student is fine-tuned on those reports under a prompt with no abnormality condition. At inference it decodes with a single forward pass, with no condition and no contrast, so the correction lives in the weights rather than in the decoding rule.

\paragraph{Contributions.}
\begin{itemize}
\item We measure abnormality information along the whole 3D CT reporting pipeline on one ruler and localise the information loss to the point where the hidden state becomes text, on two CT visual encoders and two language models.
\item We propose \textbf{SelfCue}, which narrows the surfacing gap by contrasting the generator's output distribution with and without an abnormality condition. It reaches $0.481$ macro F1@0.5 and $0.510$ GREEN on COLIPRI with Llama-3.1-8B.
\item We distil SelfCue into \textbf{SelfCue-KD}, which decodes with a single forward pass. It keeps most of the performance while no longer depending on the abnormality condition.
\end{itemize}

\section{Related work}
\label{sec:related}

\subsection{3D CT report generation}

Work on CT-RATE \citep{Hamamci_2026} has advanced largely through the visual pathway. Tokenizers built for volumes \citep{li2025mu2tokenizer} and stronger visual backbones \citep{uvlm2026} change what the language model is given, as does a cross-attention decoder built for 3D volumes \citep{hamamci2024ct2rep}. SliceWorld trains a world-state model that predicts the next slice and reads the report off the learned dynamics \citep{sliceworld2026}. Others leave the generator alone and retrieve similar studies to supply what the volume does not \citep{liang2026adaptive}. Closest to us are the decoders that put an abnormality classifier's output into the prompt \citep{10.1609/aaai.v38i3.28038,10.1609/aaai.v39i24.34788}, and CT-AGRG, which generates one sentence for each abnormality its own head predicts \citep{10981073}. SelfCue starts from the same prediction and gets more out of it, changing only the decoding rule.

\subsection{Contrastive decoding and representation intervention}

\citet{burns2023ccs} recover from a language model's activations the truth of statements its own text gets wrong. \citet{ICLR2025_a712d461} and \citet{goel2026knowingsaying} report the same gap for the model's own errors. We find the same gap in CT report generation, and contrastive decoding is the family of rules built to close it. The original form contrasts a large model with a small one and keeps only the tokens the large model still finds plausible \citep{li-etal-2023-contrastive}. Later work changes what the second branch is conditioned on: shallow layers against deep ones \citep{ICLR2024_edc36117}, an image-conditioned branch against a distorted one \citep{10657718}, a prompt against its absence \citep{pmlr-v235-sanchez24a,shi-etal-2024-trusting}. In chest radiography the rule appears anchored on an external classifier \citep{zhang-etal-2026-ccd} or on category-wise visual prompts \citep{pmlr-v315-srivastava26a}. SelfCue contrasts a prompt with its absence and closes part of the gap. Reshaping the residual stream is the other way to act on what a probe finds \citep{NEURIPS2023_81b83900}, and a recent report on medical failure regimes finds the information decodable there but not corrected by fixed linear steering \citep{decodablenotcorrected2026}. SelfCue leaves the states alone and changes only the output distribution.

\subsection{Knowledge distillation}

Removing a two-branch decoder by distilling it is a mature line in diffusion \citep{10204696}. The same has been done for autoregressive image tokens \citep{ICLR2025_a4a697a6}, and the medical analogue distils in-context vectors into a report generator \citep{dive2026}. Those students keep the condition at inference. Ours does not receive one, so the teacher holds information the student can never be given, the setting of learning under privileged information \citep{JMLR:v16:vapnik15b}. Distilling a prompt into the weights so that it need not be supplied at inference is context distillation \citep{snell2022context}. Ours distils a decoding rule rather than a prompt, and the condition the teacher reads cannot be given to the student at all. SelfCue-KD follows sequence-level distillation \citep{kim-rush-2016-sequence} rather than matching token distributions \citep{hinton2015distilling}.

\section{Surfacing gap}
\label{sec:gap}

\subsection{Probing every stage}
\label{sec:gap_ruler}

We first ask where abnormality information is retained and where it fails to reach the report. We score every stage of the pipeline on the same abnormality-level ruler. Each one is scored on the same 18 CT-RATE abnormalities, and every number is a macro average over them. The ruler is AUROC \citep{hanley1982auc, FAWCETT2006861}, which depends only on how the volumes are ranked and so is unchanged by any monotone transformation of the scores. We use the same CheapCT probe head \citep{liang2026cheapprobespredictexpensive} to score the visual tokens the encoder hands to the projector and the language model's hidden states. As a report is text, RadBERT \citep{yan2022radbert} first assigns it a probability per abnormality, and the AUROC is taken over those. We score the baseline generator's hidden states at every second layer under the same text prompt for every volume. At each layer we average the hidden states over all visual and text prompt positions, a choice Appendix~\ref{app:probe} compares against two others. The probe is fit on the training split and every number reported here is on CT-RATE validation.

\subsection{The gap along the pipeline}
\label{sec:gap_probe}

Figure~\ref{fig:overview}A traces the macro AUROC along the pipeline. Every point is measured here except COLIPRI's embedding before ORCA's compression, which is quoted from ORCA \citep{liang2026orca}. The raw volume scores $1.0$ by assumption, and the encoder carries part of that information into its embedding, where the score drops to $0.848$ on COLIPRI. The language model holds it at every layer we probe, $0.839$ to $0.848$. The report then scores $0.739$. That surfacing gap of $0.109$ opens when the hidden state becomes tokens. Swapping the visual encoder does not change this: on VISD+CT-CLIP \citep{Cao_2025_ICCV, Hamamci_2026, liang2026adaptive} the gap is $0.100$. Swapping the language model does not change it either: on MedGemma-27B \citep{sellergren2025medgemma} the gap is $0.112$, with its $62$ layers rescaled onto the $32$-layer axis of Llama-3.1-8B.

\section{Method}
\label{sec:method}

SelfCue surfaces abnormalities that a probe can recover from a CT report generator's internal representations but that are missing from its reports. It renders the probe's prediction as a textual abnormality condition and uses that condition to drive contrastive decoding. Distilling the generated reports then gives SelfCue-KD, which needs only one forward pass per token.

\subsection{Self-cued decoding}
\label{sec:selfcue}

We apply a contrastive decoding rule \citep{pmlr-v235-sanchez24a,shi-etal-2024-trusting} to narrow the surfacing gap. Let $x$ denote a chest CT volume and $M_{\mathrm{base}}$ a baseline 3D CT report generator. We threshold an abnormality readout and render the predicted abnormalities as a short sentence. That sentence is the abnormality condition, and it carries noisy information rather than ground truth. The readout can come from the hidden states or from the visual embedding. The baseline has never seen a conditional prompt, so feeding it one would be out of distribution. We therefore train $M_{\mathrm{cond}}$, warm-started from $M_{\mathrm{base}}$ and fine-tuned on CT--report pairs with the condition prepended. Its training set comes in three kinds of prompt. The first carries no condition. The second carries the ground-truth abnormalities. The third carries the same labels degraded to the probe's measured accuracy. The first keeps the plain branch in distribution. The other two show the condition at its best and at its real accuracy (Figure~\ref{fig:overview}D).

At inference the abnormality condition for this volume is prepended to its plain prompt, giving the conditional prompt (Figure~\ref{fig:overview}B). Decoding then proceeds one token at a time. At each step both branches of $M_{\mathrm{cond}}$ see the same partial report, the conditional branch under the conditional prompt and the plain branch under the plain prompt. Writing $\boldsymbol{\ell}^{\mathrm{cond}}$ and $\boldsymbol{\ell}^{\mathrm{plain}}$ for their next-token logits, SelfCue decodes from
\begin{equation}
\widetilde{\boldsymbol{\ell}}
=
\boldsymbol{\ell}^{\mathrm{cond}}
+\alpha
\left(
\boldsymbol{\ell}^{\mathrm{cond}}
-\boldsymbol{\ell}^{\mathrm{plain}}
\right),
\label{eq:selfcue}
\end{equation}
where $\alpha$ controls the contrast strength and is set to $0$ on volumes whose condition mentions no abnormality. Here that means the readout found none, not that the volume is normal. The combination promotes tokens the conditional branch favours more strongly and demotes those the plain branch favours more strongly. We greedily select the next token from $\widetilde{\boldsymbol{\ell}}$ and append it to both branches, so the two always continue from the same text.

\subsection{SelfCue-KD}
\label{sec:selfcue_kd}

SelfCue runs two forward passes of $M_{\mathrm{cond}}$ at every step. We remove this inference-time overhead through sequence-level knowledge distillation \citep{kim-rush-2016-sequence}, with SelfCue as the teacher $M_{\mathrm{teacher}}$ (Figure~\ref{fig:overview}C). Because the student receives no condition at inference, sequence-level distillation here has to meet two conditions it ordinarily does not. First, the conditions used to write the teacher corpus must be honest out of sample \citep{dao2021knowledge}. The probe is fit on the training split, so applying it back to that split returns conditions close to the ground truth and the teacher would carry information the student can never obtain. We split the training volumes into five disjoint folds $S_1,\dots,S_5$ and fit a separate probe on each set of four, so every training volume is paired with a report whose condition came from a probe that never saw it. Writing $M_{\mathrm{teacher}}^{(-k)}$ for SelfCue driven by the probe fit on the four folds other than $k$, the teacher corpus is
\begin{equation}
\mathcal{D}^{T}
=
\bigcup_{k=1}^{5}
\left\{
\left(
x_i,\; M_{\mathrm{teacher}}^{(-k)}(x_i)
\right)
:\;
x_i \in S_k
\right\}.
\label{eq:crossfit}
\end{equation}

Second, the teacher is told which abnormalities to mention and the student is not, so the student has to infer them from the image. $M_{\mathrm{student}}$ is initialized from $M_{\mathrm{cond}}$ but receives only the plain prompt. Rather than matching the teacher's token-level distributions, which would run both branches of the teacher at every training step, the student uses each complete teacher report as a sequence-level target and is optimized with standard autoregressive cross-entropy:
\begin{equation}
\mathcal{L}_{\mathrm{SeqKD}}
=
-\sum_{(x_i, y_i) \in \mathcal{D}^{T}}
\log p_{\mathrm{student}}
\left(
y_i \mid x_i
\right).
\label{eq:seqkd}
\end{equation}

At inference $M_{\mathrm{student}}$ receives only the CT volume and the plain prompt, and generates in a single pass.

\begin{table}[t]
\centering
\scriptsize
\setlength{\tabcolsep}{5pt}
\begin{tabular}{lcccccccc}
\toprule
System & F1@0.5 & $\mathrm{F1_{max}}$ & AUROC & GREEN & BLEU-1 & BLEU-4 & ROUGE-L & METEOR \\
\midrule
\multicolumn{9}{l}{\emph{Eligible works}} \\
CT-CHAT \citep{Hamamci_2026} & $0.184$ & -- & -- & -- & $0.395$ & -- & $0.334$ & -- \\
RadAgent \citep{radagent2026} & $0.220$ & -- & -- & -- & -- & -- & -- & -- \\
BTB3D \citep{NEURIPS2025_c4861cc8} & $0.258$ & -- & -- & -- & $0.439$ & $0.213$ & -- & $0.223$ \\
OmniCT \citep{ICLR2026_45419fc1} & $0.363$ & -- & -- & -- & -- & -- & -- & -- \\
U-VLM \citep{uvlm2026} & $0.414$ & -- & -- & -- & $0.474$ & $0.256$ & -- & -- \\
AdaRAG-CT \citep{liang2026adaptive} & $0.480$ & -- & -- & -- & $0.496$ & $0.242$ & $0.354$ & $0.246$ \\
CLarGen \citep{clargen2026} & $0.486$ & -- & -- & -- & -- & $0.208$ & $0.298$ & $0.335$ \\
CT-AGRG \citep{10981073} & -- & $0.501$ & -- & -- & -- & $0.172$ & $0.280$ & $0.196$ \\
$\mu^2$Tokenizer \citep{li2025mu2tokenizer} & -- & -- & -- & $0.429$ & -- & -- & -- & $0.359$ \\
SliceWorld \citep{sliceworld2026} & -- & -- & -- & $0.507$ & $0.394$ & $0.226$ & $0.382$ & $0.230$ \\
\midrule
\multicolumn{9}{l}{\emph{COLIPRI $+$ Llama-3.1-8B}} \\
$M_{\mathrm{base}}$ & $0.395$ & $0.482$ & $0.739$ & $0.466$ & $0.456$ & $0.226$ & $0.345$ & $0.214$ \\
$M_{\mathrm{cond}}$ & $0.430$ & $0.514$ & $0.762$ & $0.510$ & $0.425$ & $0.220$ & $0.356$ & $0.209$ \\
\rowcolor{blue!8}SelfCue & $0.481$ & $0.542$ & $0.775$ & $0.510$ & $0.424$ & $0.221$ & $0.356$ & $0.210$ \\
\rowcolor{green!10}SelfCue-KD & $0.462$ & $0.534$ & $0.768$ & $0.510$ & $0.416$ & $0.217$ & $0.353$ & $0.208$ \\
\midrule
\multicolumn{9}{l}{\emph{VISD$+$CT-CLIP $+$ Llama-3.1-8B}} \\
$M_{\mathrm{base}}$ & $0.408$ & $0.474$ & $0.745$ & $0.463$ & $0.461$ & $0.223$ & $0.337$ & $0.214$ \\
$M_{\mathrm{cond}}$ & $0.449$ & $0.515$ & $0.761$ & $0.512$ & $0.428$ & $0.224$ & $0.358$ & $0.211$ \\
\rowcolor{blue!8}SelfCue & $0.487$ & $0.539$ & $0.774$ & $0.504$ & $0.428$ & $0.222$ & $0.355$ & $0.211$ \\
\rowcolor{green!10}SelfCue-KD & $0.460$ & $0.526$ & $0.754$ & $0.512$ & $0.412$ & $0.218$ & $0.358$ & $0.209$ \\
\midrule
\multicolumn{9}{l}{\emph{COLIPRI $+$ MedGemma-27B}} \\
$M_{\mathrm{base}}$ & $0.348$ & $0.471$ & $0.740$ & $0.459$ & $0.451$ & $0.213$ & $0.326$ & $0.206$ \\
$M_{\mathrm{cond}}$ & $0.438$ & $0.518$ & $0.763$ & $0.472$ & $0.386$ & $0.162$ & $0.274$ & $0.201$ \\
\rowcolor{blue!8}SelfCue & $0.481$ & $0.548$ & $0.778$ & $0.456$ & $0.392$ & $0.165$ & $0.276$ & $0.204$ \\
\rowcolor{green!10}SelfCue-KD & $0.472$ & $0.535$ & $0.771$ & $0.467$ & $0.386$ & $0.160$ & $0.270$ & $0.200$ \\
\bottomrule
\end{tabular}
\caption{\textbf{Comparison on CT-RATE.} SelfCue decodes at $\alpha = 1.0$ and $\tau = 0.75$ on all three encoder--backbone pairs. $M_{\mathrm{cond}}$ is decoded with the condition in its prompt and no contrast.}
\label{tab:main}
\end{table}

\section{Experiments}
\label{sec:results}

\subsection{Setup}
\label{sec:setup}

\textbf{Dataset.} All experiments use CT-RATE \citep{Hamamci_2026} and its 18 abnormality labels. COLIPRI is scored on $n = 1564$ volumes and VISD+CT-CLIP on $n = 2987$. COLIPRI keeps one reconstruction of each scan, and reconstructions of a scan share a report. Its $1564$ volumes therefore cover every report in the validation split, so the two remain comparable.

\textbf{Encoders.} We use two visual embeddings. The first comes from ORCA \citep{liang2026orca}, which compresses the COLIPRI encoder \citep{wald2025colipri} to $216$ tokens and provides them. The second is the pairing of ViSD organ embeddings \citep{Cao_2025_ICCV} with a mean-pooled CT-CLIP embedding \citep{Hamamci_2026} introduced by \citet{liang2026adaptive}, the encoder's own $5$ tokens.

\textbf{Backbones.} The generator is a LLaVA-style pipeline \citep{NEURIPS2023_6dcf277e}: the visual tokens pass through a projector into a language model fine-tuned with LoRA \citep{hu2022lora}. We use Llama-3.1-8B-Instruct \citep{grattafiori2024llama3herdmodels} which has $32$ transformer layers, and MedGemma-27B \citep{sellergren2025medgemma}, which has $62$ layers.

\textbf{Metrics.} Abnormalities are read out of the generated text with RadBERT \citep{yan2022radbert}, as in Section~\ref{sec:gap_ruler}. We report macro AUROC, which measures how well the volumes are ranked. Alongside the widely used clinical efficacy F1, macro over the 18 abnormalities at the fixed threshold of $0.5$, we report macro $\mathrm{F1_{max}}$, the best F1 over all thresholds and therefore an upper bound. Because it maximises over thresholds, $\mathrm{F1_{max}}$ is unchanged by a systematic shift in the scores, as macro AUROC is. GREEN \citep{ostmeier-etal-2024-green} gives a language-model judgement of the report. We also report BLEU-1 and BLEU-4 \citep{papineni-etal-2002-bleu}, ROUGE-L \citep{lin-2004-rouge} and METEOR \citep{banerjee-lavie-2005-meteor}.

\textbf{Comparison eligibility.} We apply several eligibility conditions so that the systems can be compared fairly. A system must evaluate on the official CT-RATE validation split and score against the whole report, findings and impression together. It must report at least one of clinical F1 or GREEN. It must read abnormalities with the official released RadBERT weights and score with the official GREEN weights. We report the numbers each paper published itself. $10$ works pass this screen and are included in our comparison.

\textbf{Implementation.} The prompt mixture that trains $M_{\mathrm{cond}}$ is $30\%$ plain, $35\%$ ground-truth and $35\%$ ground-truth with the probe's own error rates applied. SelfCue has two hyperparameters, the contrast strength $\alpha$ and the threshold $\tau$ at which the readout becomes a condition, both chosen on the validation split at $\alpha = 1.0$ and $\tau = 0.75$. We train the projector and a rank-$128$ LoRA adapter on the language model, with the visual encoder and the compressor frozen. Optimisation uses AdamW \citep{loshchilov2019decoupled} at a learning rate of $2\times 10^{-5}$ with a cosine schedule and an effective batch size of $16$. The MedGemma baseline uses $5\times 10^{-6}$, which a sweep selected for it. Every model is trained and run on a single NVIDIA B200 GPU.

\subsection{Main results}
\label{sec:main}

BLEU, ROUGE-L and METEOR reflect fluency rather than whether the abnormalities are right \citep{liang2026cheapprobespredictexpensive}. We report them for comparability and read clinical quality from F1 and GREEN.

\textbf{Clinical efficacy.} SelfCue improves clinical efficacy on all three encoder--backbone pairs in Table~\ref{tab:main}. $M_{\mathrm{cond}}$ already improves on $M_{\mathrm{base}}$ in F1@0.5, $\mathrm{F1_{max}}$ and AUROC, and SelfCue improves on $M_{\mathrm{cond}}$ in each of them. On COLIPRI with Llama-3.1-8B, F1@0.5 rises from $0.395$ to $0.430$ and then to $0.481$, and $\mathrm{F1_{max}}$ from $0.482$ to $0.514$ and then to $0.542$. $M_{\mathrm{cond}}$ and SelfCue use the same model and the same condition and differ only in how they decode, so the gain from $M_{\mathrm{cond}}$ to SelfCue comes from our decoding rule.

The two F1 scores measure different things. $\mathrm{F1_{max}}$ takes the best threshold for each abnormality and so reflects how well the reports rank the volumes, while F1@0.5 scores the reports as they are written. Precision and recall in Appendix~\ref{app:detail} show how F1@0.5 improves. On COLIPRI recall rises from $0.365$ to $0.452$ while precision moves from $0.590$ to $0.577$. On VISD+CT-CLIP and MedGemma-27B recall rises by $0.065$ and $0.105$ with precision within $0.006$. SelfCue mentions abnormalities that $M_{\mathrm{cond}}$ left out at nearly the same precision. Appendix~\ref{app:cases} follows one volume through the four systems. On COLIPRI a report mentions $2.4$ of the $3.4$ abnormalities present, so there is room above what it reaches here. SelfCue recovers $59\%$, $61\%$ and $64\%$ of the $\mathrm{F1_{max}}$ gap between the report of $M_{\mathrm{base}}$ and the probe that reads its hidden state. It adds the most where the report leaves out the most. Sorted by prevalence, the 18 abnormalities fall into three groups of six. Against the report of $M_{\mathrm{base}}$, $\mathrm{F1_{max}}$ rises from $0.359$ to $0.463$ on the rarest group and from $0.598$ to $0.635$ on the most common. The rarest group is also where the probe exceeds the report of $M_{\mathrm{base}}$ by the widest margin.

\textbf{GREEN.} SelfCue holds GREEN at the level of $M_{\mathrm{cond}}$ on all three pairs. A higher $\tau$ raises GREEN at the cost of clinical efficacy. GREEN weighs every statement equally, and its judge counts about ten findings per report against the $3.43$ of the 18 abnormalities present in a volume. SelfCue moves a fraction of those, mentioning $2.40$ abnormalities per volume against $M_{\mathrm{cond}}$'s $2.01$, so a large gain in F1 reaches only a small part of what GREEN scores.

\textbf{SelfCue-KD.} SelfCue-KD is trained on the reports SelfCue writes and decodes with a single forward pass under the plain prompt. On COLIPRI with Llama-3.1-8B it reaches $0.462$ F1@0.5 and $0.534$ $\mathrm{F1_{max}}$, within $0.020$ and $0.008$ of its teacher, and its GREEN is the same. On VISD+CT-CLIP with Llama-3.1-8B the gap to its teacher is wider, $0.028$ and $0.014$, yet it stays above $M_{\mathrm{cond}}$ on both and ends $0.007$ ahead of its teacher on GREEN. On COLIPRI with MedGemma-27B it is within $0.009$ and $0.013$ of its teacher, stays above $M_{\mathrm{cond}}$ on both, and is $0.011$ ahead of its teacher on GREEN. It therefore runs unchanged in a standard inference engine. On one B200, SelfCue decoding one report at a time takes $6.3$ hours for the $1564$ validation volumes, while SelfCue-KD under vLLM \citep{10.1145/3600006.3613165} writes all of them in $71$ seconds.

\textbf{Eligible works.} With Llama-3.1-8B, SelfCue is on par with the strongest eligible works on F1@0.5 and GREEN and above them on $\mathrm{F1_{max}}$. CLarGen \citep{clargen2026} reaches an F1@0.5 of $0.472$ with Llama-3.1-8B as its writer, against $0.481$ for SelfCue. With MedGemma-27B it reports $0.486$ against $0.481$ for SelfCue on the same backbone. SelfCue thus depends less on the backbone than CLarGen does. SliceWorld \citep{sliceworld2026} reports a GREEN of $0.507$, against $0.510$ for both SelfCue and SelfCue-KD on COLIPRI. It gets there with a world model pretrained on DeepLesion, a second CT dataset, whereas both of ours are trained on CT-RATE alone. CT-AGRG \citep{10981073} is the only eligible work that reports $\mathrm{F1_{max}}$, $0.501$, against $0.542$ for SelfCue and $0.534$ for SelfCue-KD. The other eligible works score below SelfCue on every clinical metric they report.

\subsection{Ablations}
\label{sec:ablation}

\subsubsection{Contrast strength and condition threshold}
\label{sec:knobs}

\begin{figure}[t]
\centering
\begin{minipage}[t]{0.27\linewidth}\centering {\scriptsize $\mathrm{F1_{max}}$}\\[1pt]
\includegraphics[width=\linewidth]{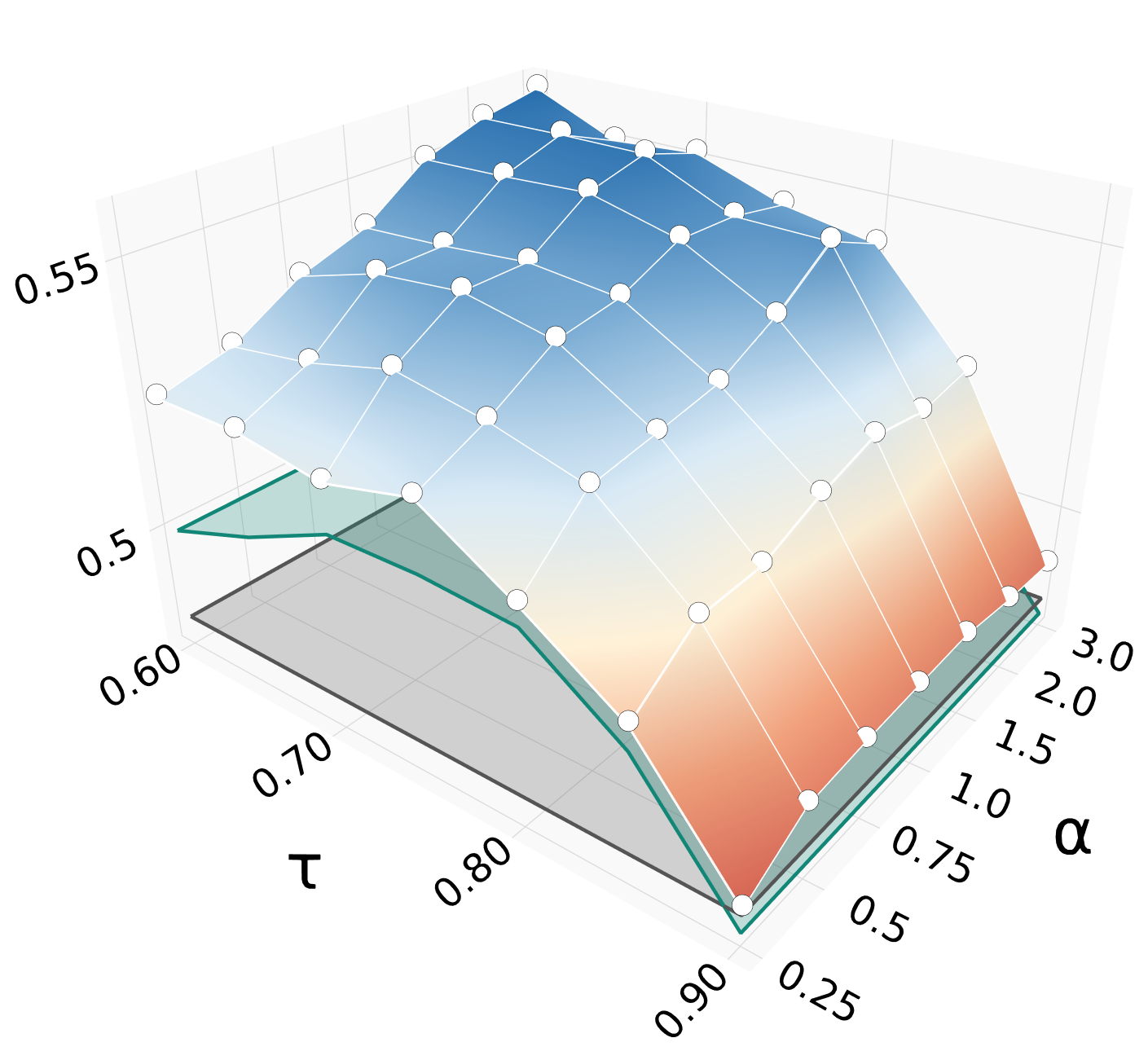}
\end{minipage}\hspace{0.05\linewidth}%
\begin{minipage}[t]{0.27\linewidth}\centering {\scriptsize F1@0.5}\\[1pt]
\includegraphics[width=\linewidth]{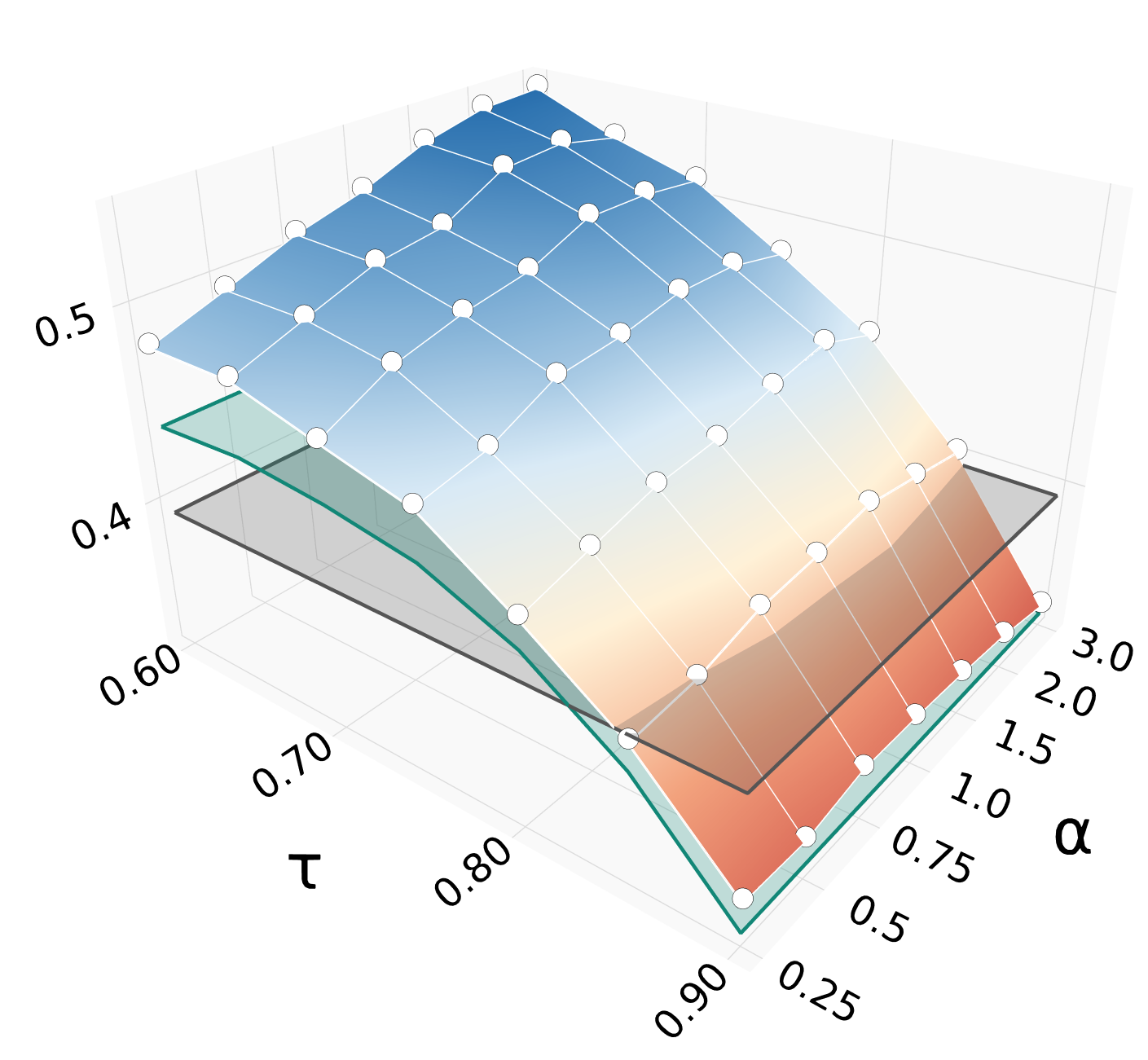}
\end{minipage}\hspace{0.05\linewidth}%
\begin{minipage}[t]{0.27\linewidth}\centering {\scriptsize GREEN}\\[1pt]
\includegraphics[width=\linewidth]{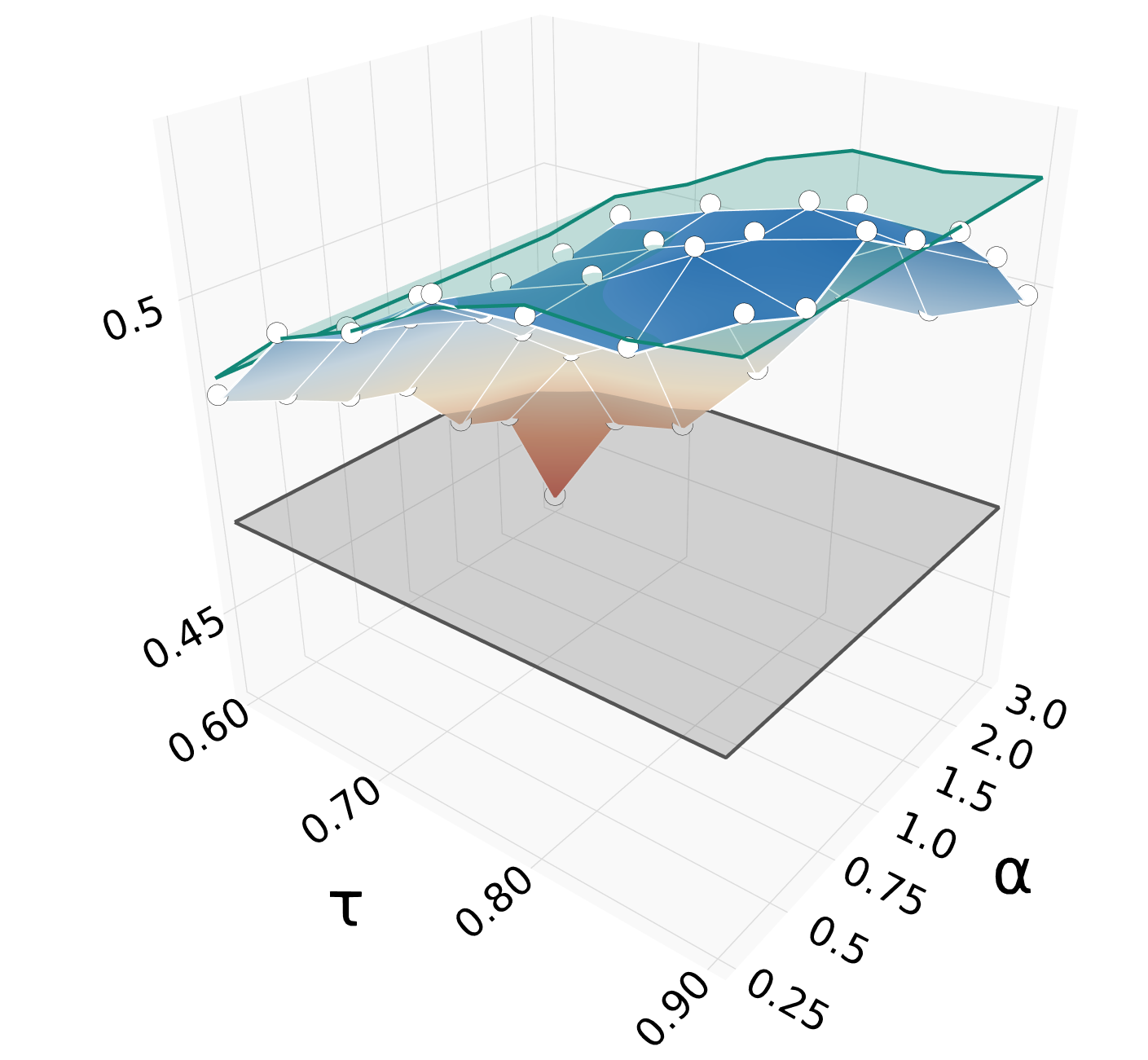}
\end{minipage}\\[2pt]
\includegraphics[width=0.36\linewidth]{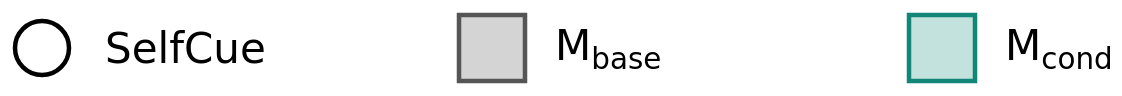}
\caption{SelfCue with the gate on empty conditions over $\alpha$ and $\tau$ on COLIPRI with Llama-3.1-8B. $M_{\mathrm{cond}}$ is decoded with the condition at the same $\tau$ and no contrast.}
\label{fig:grid3d}
\end{figure}

SelfCue has two hyperparameters. The contrast strength $\alpha$ sets how far the decoding moves away from the plain branch, and the threshold $\tau$ sets how many abnormalities the condition mentions. Figure~\ref{fig:grid3d} sweeps both on COLIPRI with Llama-3.1-8B, and Appendix~\ref{app:hpsweep} sweeps the other two pairs. At every one of the $28$ cells, SelfCue exceeds on $\mathrm{F1_{max}}$ both $M_{\mathrm{base}}$ and $M_{\mathrm{cond}}$ decoded with the same condition, and it exceeds $M_{\mathrm{cond}}$ on F1@0.5 as well, so the gain holds wherever the two are set. The threshold $\tau$ moves $\mathrm{F1_{max}}$ most. $\mathrm{F1_{max}}$ stays flat up to $\tau = 0.75$ and falls beyond it, and $M_{\mathrm{cond}}$ falls with it. A higher threshold puts fewer abnormalities in the condition, and at $\tau = 0.90$ the condition mentions none on $55\%$ of volumes. F1@0.5 falls over the whole range instead. A lower threshold makes the reports mention each abnormality more often. F1@0.5 counts these mentions at a fixed cut of $0.5$. $\mathrm{F1_{max}}$ picks the best cut for each abnormality and so barely moves.

At $\alpha = 0$, SelfCue reduces to $M_{\mathrm{cond}}$. Every positive $\alpha$ lifts $\mathrm{F1_{max}}$ above it, so the contrast itself helps. F1@0.5 follows the same course. GREEN moves the other way. At $\tau = 0.75$ it stays at the level of $M_{\mathrm{cond}}$ up to $\alpha = 1.0$ and drops at larger $\alpha$, and it rises with $\tau$. At $\tau = 0.60$ the contrast costs GREEN rather than gaining it, $0.475$ against $M_{\mathrm{cond}}$'s $0.491$. Because a higher threshold keeps only the abnormalities the probe is most confident about, the reports contain fewer false findings, which GREEN counts as errors \citep{ostmeier-etal-2024-green}.

\begin{table}[bh]
\centering
\scriptsize
\begin{tabular}{lccccccccc}
\toprule
 & \multicolumn{3}{c}{$\tau = 0.70$} & \multicolumn{3}{c}{$\tau = 0.75$} & \multicolumn{3}{c}{$\tau = 0.90$} \\
\cmidrule(lr){2-4}\cmidrule(lr){5-7}\cmidrule(lr){8-10}
$\alpha$ & F1@0.5 & $\mathrm{F1_{max}}$ & GREEN & F1@0.5 & $\mathrm{F1_{max}}$ & GREEN & F1@0.5 & $\mathrm{F1_{max}}$ & GREEN \\
\midrule
$0.25$ & $0.000$ & $0.000$ & $-0.001$ & $0.000$ & $0.000$ & $0.000$ & $+0.001$ & $+0.002$ & $-0.002$ \\
$1.0$ & $0.000$ & $0.000$ & $+0.002$ & $0.000$ & $0.000$ & $0.000$ & $+0.004$ & $+0.003$ & $+0.003$ \\
$2.0$ & $+0.003$ & $+0.003$ & $+0.022$ & $+0.004$ & $+0.004$ & $+0.029$ & $+0.008$ & $+0.009$ & $+0.038$ \\
$3.0$ & $+0.007$ & $+0.006$ & $+0.057$ & $+0.007$ & $+0.005$ & $+0.070$ & $+0.015$ & $+0.012$ & $+0.097$ \\
\bottomrule
\end{tabular}
\caption{\textbf{Gate on empty conditions.} It sets $\alpha$ to $0$ on volumes whose condition is empty, so they are decoded without contrast. Cells are gated minus ungated SelfCue on COLIPRI with Llama-3.1-8B.}
\label{tab:gate}
\end{table}

\subsubsection{Gate on empty conditions}
\label{sec:gate}

When the condition mentions no abnormality, SelfCue sets $\alpha$ to $0$ on that volume. Without this gate the contrast would amplify the sentence \emph{no significant abnormality} against a plain branch that carries no condition at all. Table~\ref{tab:gate} compares gated and ungated decoding on COLIPRI with Llama-3.1-8B. The gate leaves F1@0.5 and $\mathrm{F1_{max}}$ nearly unchanged everywhere. Its effect on GREEN grows with $\alpha$, and with $\tau$, since a higher threshold leaves more conditions empty. How much the gate matters depends on the pair. On COLIPRI with MedGemma-27B it already raises GREEN by $0.027$ at the operating point.

\subsubsection{Condition source}
\label{sec:condsource}

The abnormality condition is used in three places, to train $M_{\mathrm{cond}}$, to decode with SelfCue and to write the teacher corpus. We implement two condition sources, a probe on the LLM hidden states and a probe on the visual embedding. On the training set a probe fit on that set cannot be applied to it directly, so the conditions there are either ground truth with injected probe-level noise or 5-fold cross-fitted predictions. We read the probe off the hidden states, train $M_{\mathrm{cond}}$ under injected noise and write the teacher corpus under cross-fitted conditions. Table~\ref{tab:condsource} swaps each choice for the other option, one at a time. Reading the probe off the visual embedding instead changes every measure by at most $0.008$. How the training set conditions are produced matters more, and in opposite directions for the two uses. Training $M_{\mathrm{cond}}$ on cross-fitted conditions lowers its $\mathrm{F1_{max}}$ by $0.021$. Writing the teacher corpus under injected noise lowers the student's $\mathrm{F1_{max}}$ by $0.063$, and the student mentions $1.53$ abnormalities per volume instead of $2.25$. The student sees only the image. Injected noise is independent of the volume, so nothing in the image tells the student where the condition errs, whereas cross-fitted errors fall on the volumes the probe finds hard.

\begin{table}[t]
\centering
\scriptsize
\begin{tabular}{lcccccc}
\toprule
 & \multicolumn{3}{c}{Condition source} & \multicolumn{3}{c}{Training set condition} \\
\cmidrule(lr){2-4}\cmidrule(lr){5-7}
Where it is used & F1@0.5 & $\mathrm{F1_{max}}$ & GREEN & F1@0.5 & $\mathrm{F1_{max}}$ & GREEN \\
\midrule
Training $M_{\mathrm{cond}}$ & $+0.008$ & $+0.004$ & $-0.001$ & $-0.026$ & $-0.021$ & $-0.017$ \\
SelfCue decoding & $+0.008$ & $0.000$ & $-0.003$ & -- & -- & -- \\
Teacher corpus & $+0.005$ & $-0.003$ & $-0.004$ & $-0.151$ & $-0.063$ & $-0.013$ \\
\bottomrule
\end{tabular}
\caption{\textbf{Condition source.} It is either the LLM hidden states or the visual embedding, and the training set conditions are ground truth with injected noise or 5-fold cross-fitted predictions. Each cell is the change when one choice is replaced by the other, on COLIPRI with Llama-3.1-8B.}
\label{tab:condsource}
\end{table}

\subsubsection{Sequence-level distillation}
\label{sec:seqkd}

SelfCue-KD is trained on the teacher's text. The alternative is token-level distillation \citep{hinton2015distilling}, which matches the student's next-token distribution to the teacher's at every position of the reference reports. Table~\ref{tab:seqkd} compares the two students, which start from the same checkpoint and train for the same number of steps. Appendix~\ref{app:studentinit} varies that checkpoint and the teacher reports the student sees. The two losses produce nearly the same student. The sequence-level student is ahead by $0.010$ F1@0.5 and $0.002$ $\mathrm{F1_{max}}$ and behind by $0.002$ GREEN. What the student ends up doing is set by the corpus it is trained on far more than by the loss. We use the sequence-level loss because it is ordinary fine-tuning on text and never holds the teacher in memory beside the student.

The mention rates show what the teacher writes and what the student keeps. The contrast pushes the abnormalities the condition lists into the report. Given the condition in its prompt, $M_{\mathrm{cond}}$ mentions $0.70$ of the listed abnormalities that are present, and SelfCue raises this to $0.90$. It pushes the listed abnormalities that are absent almost as far, from $0.61$ to $0.86$, so the probe's errors pass into the teacher's reports. The students inherit far less of them. They decode without the condition and read only the image, and SelfCue-KD is more selective than $M_{\mathrm{cond}}$, which reads the condition. It mentions more of the listed abnormalities that are present, $0.73$, fewer of those that are absent, $0.58$, and three times as many of the present abnormalities the condition leaves out, $0.11$ against $0.04$. So besides speed, distillation leaves the deployed model less bound to the probe.

\begin{table}[thb]
\centering
\scriptsize
\begin{tabular}{lcccccc}
\toprule
 & $M_{\mathrm{base}}$ & $M_{\mathrm{cond}}$, plain prompt & $M_{\mathrm{cond}}$ & SelfCue & SelfCue-KD & Token-level KD \\
\midrule
F1@0.5 & $0.395$ & $0.299$ & $0.430$ & $0.481$ & $0.462$ & $0.452$ \\
$\mathrm{F1_{max}}$ & $0.482$ & $0.438$ & $0.514$ & $0.542$ & $0.534$ & $0.532$ \\
AUROC & $0.739$ & $0.709$ & $0.762$ & $0.775$ & $0.768$ & $0.765$ \\
GREEN & $0.466$ & $0.495$ & $0.510$ & $0.510$ & $0.510$ & $0.512$ \\
\midrule
In condition, present & $0.505$ & $0.364$ & $0.697$ & $0.901$ & $0.729$ & $0.695$ \\
In condition, absent & $0.386$ & $0.259$ & $0.610$ & $0.855$ & $0.578$ & $0.554$ \\
Outside condition, present & $0.148$ & $0.095$ & $0.039$ & $0.018$ & $0.112$ & $0.121$ \\
Outside condition, absent & $0.055$ & $0.030$ & $0.009$ & $0.006$ & $0.019$ & $0.020$ \\
\bottomrule
\end{tabular}
\caption{\textbf{Sequence-level distillation.} The last four rows are mention rates, split by whether the condition lists the abnormality and whether it is present, averaged over the 18 abnormalities. Results are on COLIPRI with Llama-3.1-8B.}
\label{tab:seqkd}
\end{table}

\subsection{Selection on held-out data}
\label{sec:selection}

CT-RATE has no test split, so Table~\ref{tab:main} is scored on the same validation split its settings were chosen on. A held-out experiment in Table~\ref{tab:heldout} confirms that the reported setting and its gains remain consistent without the validation split. We carve $3{,}000$ training volumes by patient and train on the remaining $21{,}128$ at three seeds, with the recipe unchanged. All three seeds land on that setting. The models here see $12\%$ less data and still perform at the level of Table~\ref{tab:main}, with $M_{\mathrm{base}}$'s GREEN the one deviation. SelfCue gains $0.059$ F1@0.5 and $0.031$ $\mathrm{F1_{max}}$ over $M_{\mathrm{cond}}$, against $0.052$ and $0.028$ there. Across seeds, the standard deviation is at most $0.012$ on every measure. SelfCue therefore holds its gain both without the split its setting was chosen on and across independent training runs.

\begin{table}[t]
\centering
\scriptsize
\begin{tabular}{lcccc}
\toprule
 & F1@0.5 & $\mathrm{F1_{max}}$ & AUROC & GREEN \\
\midrule
$M_{\mathrm{base}}$ & {\color{gray}$0.395$}\enspace $0.424 \pm 0.008$ & {\color{gray}$0.482$}\enspace $0.487 \pm 0.006$ & {\color{gray}$0.739$}\enspace $0.750 \pm 0.006$ & {\color{gray}$0.466$}\enspace $0.431 \pm 0.004$ \\
$M_{\mathrm{cond}}$ & {\color{gray}$0.430$}\enspace $0.430 \pm 0.006$ & {\color{gray}$0.514$}\enspace $0.510 \pm 0.007$ & {\color{gray}$0.762$}\enspace $0.757 \pm 0.001$ & {\color{gray}$0.510$}\enspace $0.510 \pm 0.006$ \\
SelfCue & {\color{gray}$0.481$}\enspace $0.489 \pm 0.012$ & {\color{gray}$0.542$}\enspace $0.541 \pm 0.005$ & {\color{gray}$0.775$}\enspace $0.774 \pm 0.005$ & {\color{gray}$0.510$}\enspace $0.503 \pm 0.007$ \\
SelfCue-KD & {\color{gray}$0.462$}\enspace $0.463 \pm 0.001$ & {\color{gray}$0.534$}\enspace $0.529 \pm 0.002$ & {\color{gray}$0.768$}\enspace $0.763 \pm 0.002$ & {\color{gray}$0.510$}\enspace $0.510 \pm 0.008$ \\
\bottomrule
\end{tabular}
\caption{\textbf{Held-out selection.} COLIPRI with Llama-3.1-8B on the validation split. Each cell gives the mean and standard deviation over three seeds, after the reported value in grey.}
\label{tab:heldout}
\end{table}

\section{Discussion}
\label{sec:discussion}

\textbf{The limit is the readout.} SelfCue closes about $60\%$ of the distance between the report and the probe that reads the generator's hidden state, so some of what is left is still open to the decoding rule. The larger part is not. Given the ground-truth labels as its condition, the same rule reaches $0.869$ macro F1@0.5 against SelfCue's $0.480$ on COLIPRI with Llama-3.1-8B (Appendix~\ref{app:provenance}), so what a report can say is bounded by what the readout gets right.

\textbf{F1 against GREEN.} The two metrics count different things. Macro F1 weighs the 18 abnormalities equally, and SelfCue gains most on the rarest of them, the abnormalities a report is most likely to leave out. GREEN counts everything a report says, of which the 18 abnormalities are a small part, and a condition that is right about every one of them leaves it where it was. SelfCue thus raises the clinical content of a report while keeping its agreement with the reference level.

\textbf{Knowing and saying.} Earlier work reads the gap between what a model holds and what it says out of its activations \citep{burns2023ccs, ICLR2025_a712d461, goel2026knowingsaying}. We close part of it instead of measuring it, in a setting where the text is the clinical product rather than an answer to be checked. The rule needs only a readout of that state and a generator that accepts a condition, so it carries to other report generators. Our evidence comes from CT-RATE and its 18 labelled abnormalities, and other imaging modalities and reader studies are the next step.

\textbf{Future work.} What SelfCue writes into the condition is limited to the 18 abnormalities that CT-RATE labels, a small part of what GREEN counts. The decoding rule is not tied to these labels. A probe trained on richer annotations, such as where an abnormality lies, how large it is or findings outside the 18, could fill the same condition sentence. A condition of that kind is what it would take for the gain to reach GREEN as well as F1. A second direction is iterative training. SelfCue-KD already mentions three times as many of the present abnormalities that the condition leaves out as $M_{\mathrm{cond}}$ does. The student can take the place of $M_{\mathrm{base}}$, and the whole pipeline of probing, conditioning, decoding and distilling can run again on it. Whether each round adds to the last is open.

\section{Conclusion}

A 3D CT report generator holds more about the abnormalities in a volume than its report says. The information survives every layer of the model and is lost only when the hidden state becomes text. SelfCue reads it back from the generator's own hidden state and uses it to steer decoding, raising clinical efficacy, and SelfCue-KD moves that correction into the weights so the deployed model decodes in a single pass. Surfacing what a model already knows is a route to better reports that complements stronger encoders and larger datasets.

\clearpage
\bibliography{refs}
\bibliographystyle{plainnat}

\clearpage
\appendix
\setcounter{secnumdepth}{2}
\begin{center}
{\LARGE\bfseries Supplementary Material}\\[0.45em]
{\large\bfseries SelfCue: Making a 3D CT Report Generator Say What It Already Knows}
\end{center}
\vspace{0.6em}

\section{Terminology}
\label{app:terms}

This appendix collects the terms the paper uses, with examples where they help.

\noindent\textbf{Abnormality condition}. A sentence mentioning the abnormalities predicted present for this volume. The prediction is made at inference and is noisy, not a ground-truth annotation.

\promptbox{Findings to report: Atelectasis, Lung opacity, Mosaic attenuation pattern, Peribronchial thickening, Consolidation.}

\noindent\textbf{Plain prompt}. A report-generation instruction carrying no abnormality condition, drawn per volume from the 42 report-generation questions CT-RATE provides.

\promptbox{\textless image\textgreater{} Could you write the radiology report for this chest CT scan? \textless report\_generation\textgreater}

\noindent\textbf{Conditional prompt}. That volume's plain prompt with the abnormality condition prepended.

\promptbox{\textless image\textgreater{} Findings to report: Atelectasis, Lung opacity, Mosaic attenuation pattern, Peribronchial thickening, Consolidation. Could you write the radiology report for this chest CT scan? \textless report\_generation\textgreater}

\noindent$\boldsymbol{M_{\mathrm{base}}}$. The baseline generator, trained on CT volumes paired with their reports under plain prompts.

\noindent$\boldsymbol{M_{\mathrm{cond}}}$. Warm-started from $M_{\mathrm{base}}$ and trained with the abnormality condition prepended. $30\%$ of its training examples carry no condition, so the plain prompt stays in distribution.

\noindent$\boldsymbol{M_{\mathrm{teacher}}}$. The SelfCue decoder that writes the reports the student is trained on.

\noindent$\boldsymbol{M_{\mathrm{student}}}$. Trained on the teacher's reports under plain prompts. It receives no abnormality condition at inference. This model is SelfCue-KD.

\noindent\textbf{Plain branch}. The forward pass of $M_{\mathrm{cond}}$ under the plain prompt.

\noindent\textbf{Conditional branch}. The forward pass of $M_{\mathrm{cond}}$ under the conditional prompt.

\section{Measurement implementation}
\label{app:measurement}

\subsection{Probe implementation}
\label{app:probe}

Both sides of Figure~\ref{fig:overview}A are read by the same head, the CheapCT probe \citep{liang2026cheapprobespredictexpensive}. It is fit on the training split with AdamW at a learning rate of $10^{-3}$, weight decay $10^{-4}$ and batch $64$, for at most $50$ epochs with early stopping on patience $10$. Inside the language model it is fit separately on every second layer, and the layer with the highest macro AUROC is the one reported. Table~\ref{tab:pooling} reads the hidden states three ways: averaged over all visual and text prompt positions, averaged over the visual prompt alone, and taken at the last prompt position. The three agree to within $0.010$ macro AUROC at their best layers. We average over all prompt positions throughout, because causal attention lets every generated position attend to the whole prompt.

\begin{table}[ht]
\centering
\scriptsize
\begin{tabular}{lccc}
\toprule
 & all prompt positions & visual prompt only & last prompt position \\
\midrule
COLIPRI, Llama-3.1-8B & $0.848$ & $0.848$ & $0.842$ \\
VISD+CT-CLIP, Llama-3.1-8B & $0.845$ & $0.838$ & $0.835$ \\
COLIPRI, MedGemma-27B & $0.852$ & $0.851$ & $0.847$ \\
\bottomrule
\end{tabular}
\caption{\textbf{Probe pooling.} Macro AUROC over the 18 CT-RATE abnormalities at each probe's best layer.}
\label{tab:pooling}
\end{table}

\subsection{Thresholds and $\mathrm{F1_{max}}$}
\label{app:thresholds}

\begin{figure}[h]
\centering
\includegraphics[width=0.49\linewidth]{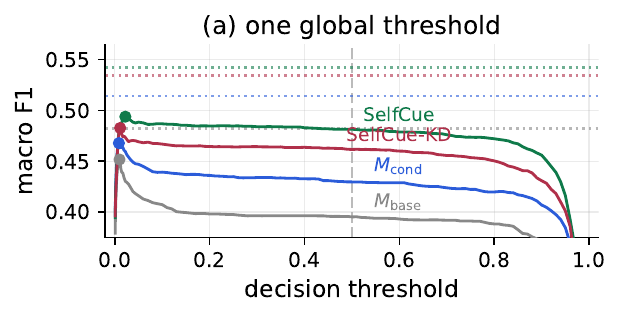}
\hfill
\includegraphics[width=0.49\linewidth]{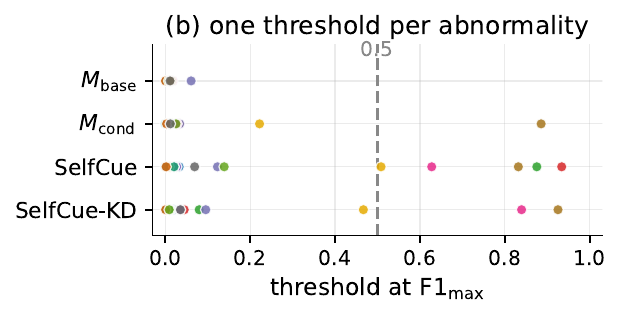}
\caption{Thresholds on RadBERT's probability for COLIPRI with Llama-3.1-8B. (a) Markers give each curve's peak and dotted lines the reported $\mathrm{F1_{max}}$. (b) One colour per abnormality.}
\label{fig:f1thr}
\end{figure}

F1@0.5 and $\mathrm{F1_{max}}$ differ only in how the cut on RadBERT's probability is chosen, and Figure~\ref{fig:f1thr} shows what that choice does. It does not decide the comparison. The order of the four systems survives every global cut between $0.01$ and $0.95$ and survives fitting a cut per abnormality. What it decides is the level. No single cut reaches $\mathrm{F1_{max}}$, which buys its extra $0.03$ to $0.05$ at cuts that differ between abnormalities and mostly sit far below $0.5$. $\mathrm{F1_{max}}$ therefore bounds how well the reports rank rather than marking an operating point, which is why F1@0.5 is reported beside it.

\section{Detailed results}
\label{app:results}

\subsection{Case study} \label{app:cases} 
\begin{figure}[!htbp]
\centering
\includegraphics[
  width=\linewidth,
  trim={0 170bp 0 70bp},
  clip
]{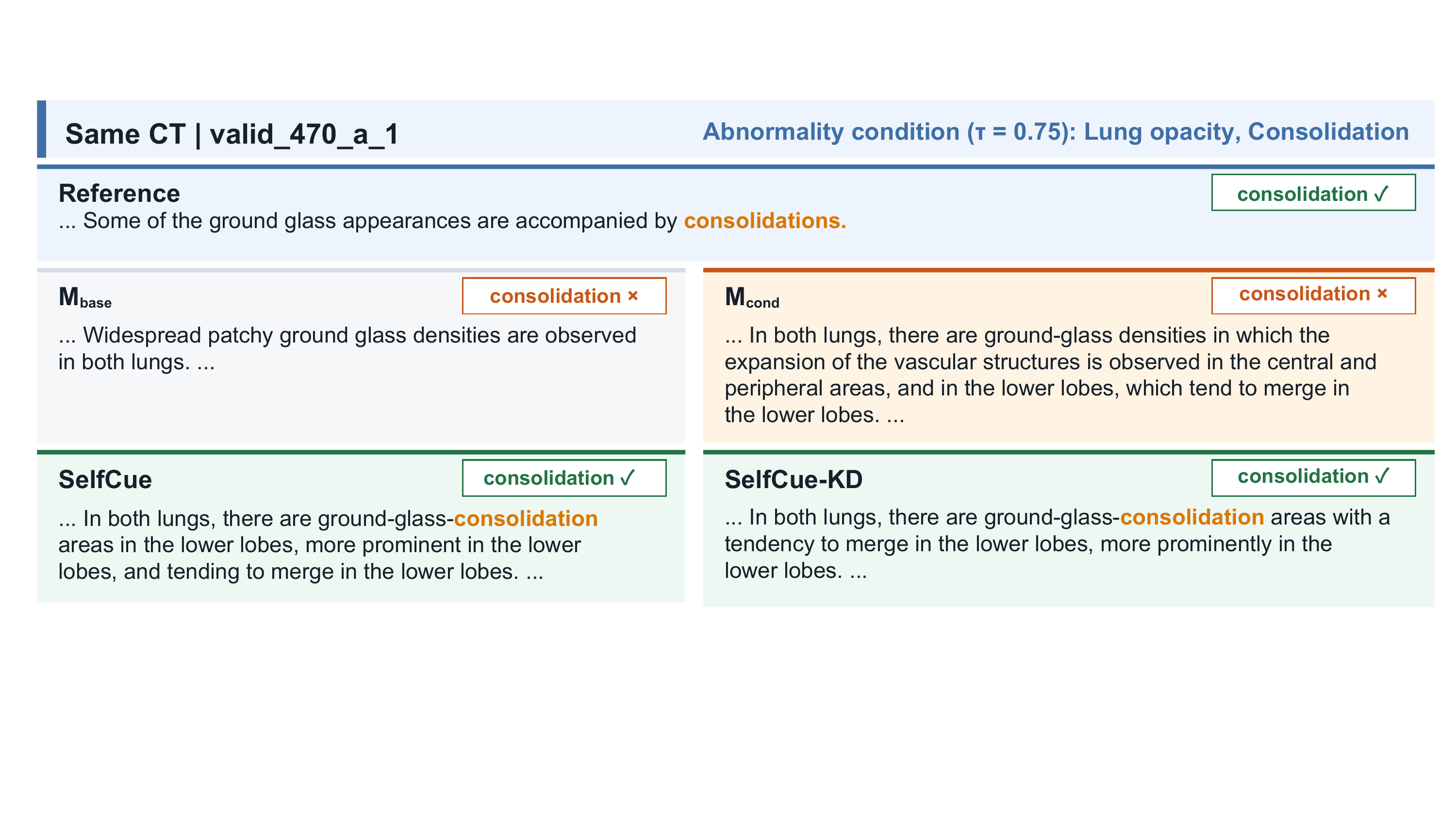}
\caption{\textbf{Case study on one CT volume.} Orange marks consolidation and the symbols show whether each report mentions it.} 
\label{fig:case_study} 
\end{figure} 
Figure~\ref{fig:case_study} shows one validation volume where consolidation is present in both the reference and the abnormality condition. $M_{\mathrm{base}}$ omits it. $M_{\mathrm{cond}}$ makes the description longer but still omits it, so the condition alone is not enough. SelfCue surfaces consolidation through contrastive decoding, and SelfCue-KD keeps it under a plain prompt. Hiatal hernia is also present in the reference, but it is absent from the condition and from all four generated reports. The correction therefore stops where the readout does.

\subsection{Training curves}
\label{app:ckpt}

\begin{figure}[h]
\centering
\includegraphics[width=\linewidth]{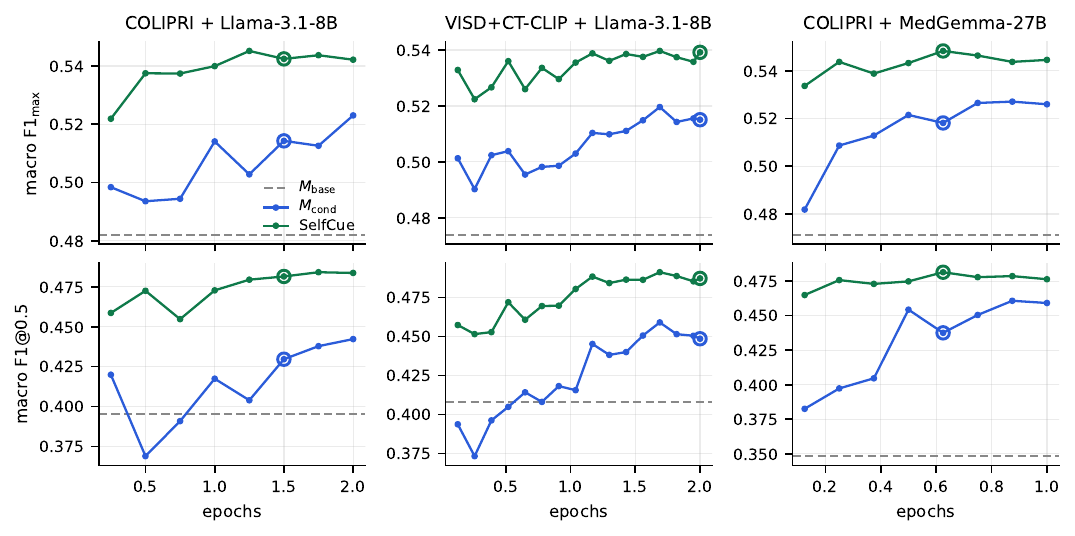}
\caption{Clinical efficacy at every checkpoint of the conditional runs. Open circles mark the checkpoint carried into every other result in the paper.}
\label{fig:ckptcurves}
\end{figure}

Every checkpoint of every model was decoded and scored, and Figure~\ref{fig:ckptcurves} shows the conditional runs. $M_{\mathrm{base}}$ is read at its last checkpoint, and $M_{\mathrm{cond}}$ and SelfCue, which decodes from it, share one checkpoint. The gain of SelfCue over $M_{\mathrm{cond}}$ is positive on F1@0.5 and $\mathrm{F1_{max}}$ at every checkpoint, though its size moves. F1@0.5 is the less stable of the two. Across the checkpoints of $M_{\mathrm{base}}$ it varies three to four times as much as $\mathrm{F1_{max}}$ on all three pairs, with a standard deviation of $0.037$ against $0.011$ on COLIPRI with Llama-3.1-8B. A fixed cut at $0.5$ tracks how often a report mentions an abnormality, which $\mathrm{F1_{max}}$ absorbs by choosing the cut for each abnormality.

\subsection{Clinical efficacy}
\label{app:detail}

Table~\ref{tab:pr} splits F1@0.5 into precision and recall. The gain of SelfCue over $M_{\mathrm{base}}$ is mostly recall, $0.338$ to $0.452$ on COLIPRI with Llama-3.1-8B, and it mentions no more abnormalities per volume than the baseline does, so it wins recall by mentioning different abnormalities rather than more of them. Its precision and recall land near the condition's own.

Table~\ref{tab:greendetail} decomposes GREEN into the counts the judge writes for each report \citep{ostmeier-etal-2024-green}. Against $M_{\mathrm{base}}$, SelfCue writes fewer false findings on all three pairs. On the two pairs with Llama-3.1-8B it also matches more and misses fewer, and GREEN rises, though the rise is already there in $M_{\mathrm{cond}}$. On COLIPRI with MedGemma-27B it matches fewer and misses more, which cancels the drop in false findings and leaves GREEN where it was. The length of a report moves with the conditional stage rather than the contrast. On COLIPRI with MedGemma-27B $M_{\mathrm{cond}}$ writes half again as many words as $M_{\mathrm{base}}$ while matching fewer findings, and SelfCue leaves the length where it found it.

We sort the 18 abnormalities by prevalence and split them into three groups of six, separated by rules in Table~\ref{tab:perfinding}. The gain of SelfCue over $M_{\mathrm{base}}$ concentrates on the rarest group, $0.105$ against $0.039$ and $0.037$ on COLIPRI with Llama-3.1-8B, and the rank correlation between prevalence and gain runs from $-0.51$ to $-0.65$ across the three pairs. The surfacing gap is widest on the rarest group too, though it separates the groups far less.

\begin{table}[h]
\centering
\scriptsize
\begin{tabular}{lccccc}
\toprule
System & F1 & $\mathrm{F1_{max}}$ & Precision & Recall & Mentioned / volume \\
\midrule
\multicolumn{6}{l}{\emph{COLIPRI $+$ Llama-3.1-8B}} \\
Condition ($\tau = 0.75$) & $0.500$ & -- & $0.574$ & $0.491$ & $2.54$ \\
$M_{\mathrm{base}}$ & $0.395$ & $0.482$ & $0.509$ & $0.338$ & $2.37$ \\
$M_{\mathrm{cond}}$ & $0.430$ & $0.514$ & $0.590$ & $0.365$ & $2.01$ \\
\rowcolor{blue!8}SelfCue & $0.481$ & $0.542$ & $0.577$ & $0.452$ & $2.40$ \\
\rowcolor{green!10}SelfCue-KD & $0.462$ & $0.534$ & $0.580$ & $0.434$ & $2.25$ \\
\midrule
\multicolumn{6}{l}{\emph{VISD$+$CT-CLIP $+$ Llama-3.1-8B}} \\
Condition ($\tau = 0.75$) & $0.503$ & -- & $0.565$ & $0.507$ & $2.74$ \\
$M_{\mathrm{base}}$ & $0.408$ & $0.474$ & $0.505$ & $0.357$ & $2.56$ \\
$M_{\mathrm{cond}}$ & $0.449$ & $0.515$ & $0.569$ & $0.398$ & $2.29$ \\
\rowcolor{blue!8}SelfCue & $0.487$ & $0.539$ & $0.563$ & $0.463$ & $2.55$ \\
\rowcolor{green!10}SelfCue-KD & $0.460$ & $0.526$ & $0.585$ & $0.412$ & $2.15$ \\
\midrule
\multicolumn{6}{l}{\emph{COLIPRI $+$ MedGemma-27B}} \\
Condition ($\tau = 0.75$) & $0.483$ & -- & $0.583$ & $0.505$ & $2.60$ \\
$M_{\mathrm{base}}$ & $0.348$ & $0.471$ & $0.529$ & $0.302$ & $2.26$ \\
$M_{\mathrm{cond}}$ & $0.438$ & $0.518$ & $0.577$ & $0.381$ & $2.14$ \\
\rowcolor{blue!8}SelfCue & $0.481$ & $0.548$ & $0.576$ & $0.485$ & $2.53$ \\
\rowcolor{green!10}SelfCue-KD & $0.472$ & $0.535$ & $0.595$ & $0.424$ & $2.21$ \\
\bottomrule
\end{tabular}
\caption{\textbf{Precision and recall} as macro averages over the 18 abnormalities. The condition row is the probe at $\tau = 0.75$. $3.43$ abnormalities are present per volume on COLIPRI and $3.47$ on VISD$+$CT-CLIP.}
\label{tab:pr}
\end{table}

\begin{table}[h]
\centering
\scriptsize
\begin{tabular}{lccccccc}
\toprule
System & $n$ & GREEN & Matched & False finding & Missing & Location & Words \\
\midrule
\multicolumn{8}{l}{\emph{COLIPRI $+$ Llama-3.1-8B}} \\
$M_{\mathrm{base}}$ & $1548$ & $0.470$ & $5.01$ & $1.47$ & $3.45$ & $0.22$ & $194$ \\
$M_{\mathrm{cond}}$ & $1555$ & $0.513$ & $5.39$ & $1.17$ & $3.15$ & $0.15$ & $177$ \\
\rowcolor{blue!8}SelfCue & $1560$ & $0.511$ & $5.36$ & $1.19$ & $3.16$ & $0.17$ & $174$ \\
\rowcolor{green!10}SelfCue-KD & $1558$ & $0.512$ & $5.63$ & $1.09$ & $3.47$ & $0.14$ & $172$ \\
\midrule
\multicolumn{8}{l}{\emph{VISD$+$CT-CLIP $+$ Llama-3.1-8B}} \\
$M_{\mathrm{base}}$ & $2956$ & $0.467$ & $5.02$ & $1.57$ & $3.37$ & $0.22$ & $200$ \\
$M_{\mathrm{cond}}$ & $2965$ & $0.515$ & $5.54$ & $1.23$ & $3.13$ & $0.16$ & $179$ \\
\rowcolor{blue!8}SelfCue & $2970$ & $0.507$ & $5.44$ & $1.27$ & $3.16$ & $0.19$ & $179$ \\
\rowcolor{green!10}SelfCue-KD & $2955$ & $0.517$ & $5.81$ & $1.10$ & $3.37$ & $0.15$ & $171$ \\
\midrule
\multicolumn{8}{l}{\emph{COLIPRI $+$ MedGemma-27B}} \\
$M_{\mathrm{base}}$ & $1548$ & $0.463$ & $4.82$ & $1.46$ & $3.56$ & $0.18$ & $199$ \\
$M_{\mathrm{cond}}$ & $1557$ & $0.474$ & $4.26$ & $0.79$ & $3.80$ & $0.11$ & $306$ \\
\rowcolor{blue!8}SelfCue & $1553$ & $0.459$ & $4.14$ & $0.83$ & $3.90$ & $0.12$ & $302$ \\
\rowcolor{green!10}SelfCue-KD & $1557$ & $0.469$ & $4.24$ & $0.76$ & $3.85$ & $0.10$ & $302$ \\
\bottomrule
\end{tabular}
\caption{\textbf{GREEN components} as mean counts per report. $n$ excludes the reports whose judge output could not be parsed.}
\label{tab:greendetail}
\end{table}

\begin{table}[h]
\centering
\scriptsize
\begin{tabular}{lccccccc}
\toprule
 & & \multicolumn{2}{c}{COLIPRI, Llama-3.1-8B} & \multicolumn{2}{c}{VISD+CT-CLIP, Llama-3.1-8B} & \multicolumn{2}{c}{COLIPRI, MedGemma-27B} \\
Abnormality & Prevalence & Gap & $\Delta\mathrm{F1_{max}}$ & Gap & $\Delta\mathrm{F1_{max}}$ & Gap & $\Delta\mathrm{F1_{max}}$ \\
\midrule
Pericardial effusion & $7.0\%$ & $0.123$ & $+0.151$ & $0.151$ & $+0.176$ & $0.121$ & $+0.073$ \\
Interlobular septal thickening & $7.9\%$ & $0.094$ & $+0.059$ & $0.055$ & $+0.071$ & $0.070$ & $+0.074$ \\
Mosaic attenuation pattern & $8.1\%$ & $0.139$ & $+0.148$ & $0.144$ & $+0.176$ & $0.176$ & $+0.165$ \\
Medical material & $10.1\%$ & $0.169$ & $+0.105$ & $0.107$ & $+0.095$ & $0.195$ & $+0.279$ \\
Cardiomegaly & $10.5\%$ & $0.085$ & $+0.090$ & $0.089$ & $+0.104$ & $0.108$ & $+0.171$ \\
Bronchiectasis & $10.8\%$ & $0.199$ & $+0.077$ & $0.111$ & $+0.060$ & $0.153$ & $+0.121$ \\
\midrule
Peribronchial thickening & $11.2\%$ & $0.100$ & $+0.021$ & $0.099$ & $+0.072$ & $0.083$ & $+0.039$ \\
Pleural effusion & $12.1\%$ & $0.071$ & $+0.064$ & $0.063$ & $+0.096$ & $0.086$ & $+0.126$ \\
Hiatal hernia & $13.9\%$ & $0.074$ & $-0.004$ & $0.135$ & $+0.005$ & $0.095$ & $+0.026$ \\
Consolidation & $19.1\%$ & $0.108$ & $+0.136$ & $0.091$ & $+0.108$ & $0.103$ & $+0.122$ \\
Emphysema & $19.5\%$ & $0.106$ & $+0.023$ & $0.101$ & $+0.018$ & $0.148$ & $+0.023$ \\
Atelectasis & $23.3\%$ & $0.111$ & $-0.006$ & $0.102$ & $-0.012$ & $0.110$ & $+0.007$ \\
\midrule
Coronary artery wall calcification & $24.8\%$ & $0.126$ & $+0.107$ & $0.108$ & $+0.082$ & $0.107$ & $+0.091$ \\
Lymphadenopathy & $25.4\%$ & $0.054$ & $+0.000$ & $0.053$ & $+0.018$ & $0.071$ & $+0.020$ \\
Pulmonary fibrotic sequela & $27.3\%$ & $0.127$ & $+0.008$ & $0.104$ & $-0.003$ & $0.108$ & $-0.034$ \\
Arterial wall calcification & $27.8\%$ & $0.112$ & $+0.110$ & $0.097$ & $+0.094$ & $0.101$ & $+0.088$ \\
Lung opacity & $38.7\%$ & $0.044$ & $+0.000$ & $0.051$ & $+0.014$ & $0.062$ & $+0.002$ \\
Lung nodule & $45.0\%$ & $0.125$ & $-0.002$ & $0.133$ & $+0.000$ & $0.119$ & $-0.005$ \\
\bottomrule
\end{tabular}
\caption{\textbf{Gap and gain per abnormality.} Both columns are read against $M_{\mathrm{base}}$, the gap as probe minus report in AUROC and the gain as SelfCue minus report in $\mathrm{F1_{max}}$. Prevalence is on the COLIPRI split.}
\label{tab:perfinding}
\end{table}

\subsection{The sweep on the other pairs}
\label{app:hpsweep}

\begin{figure}[h]
\centering
{\scriptsize VISD$+$CT-CLIP with Llama-3.1-8B}\\[1pt]
\begin{minipage}[t]{0.27\linewidth}\centering {\scriptsize $\mathrm{F1_{max}}$}\\[1pt]
\includegraphics[width=\linewidth]{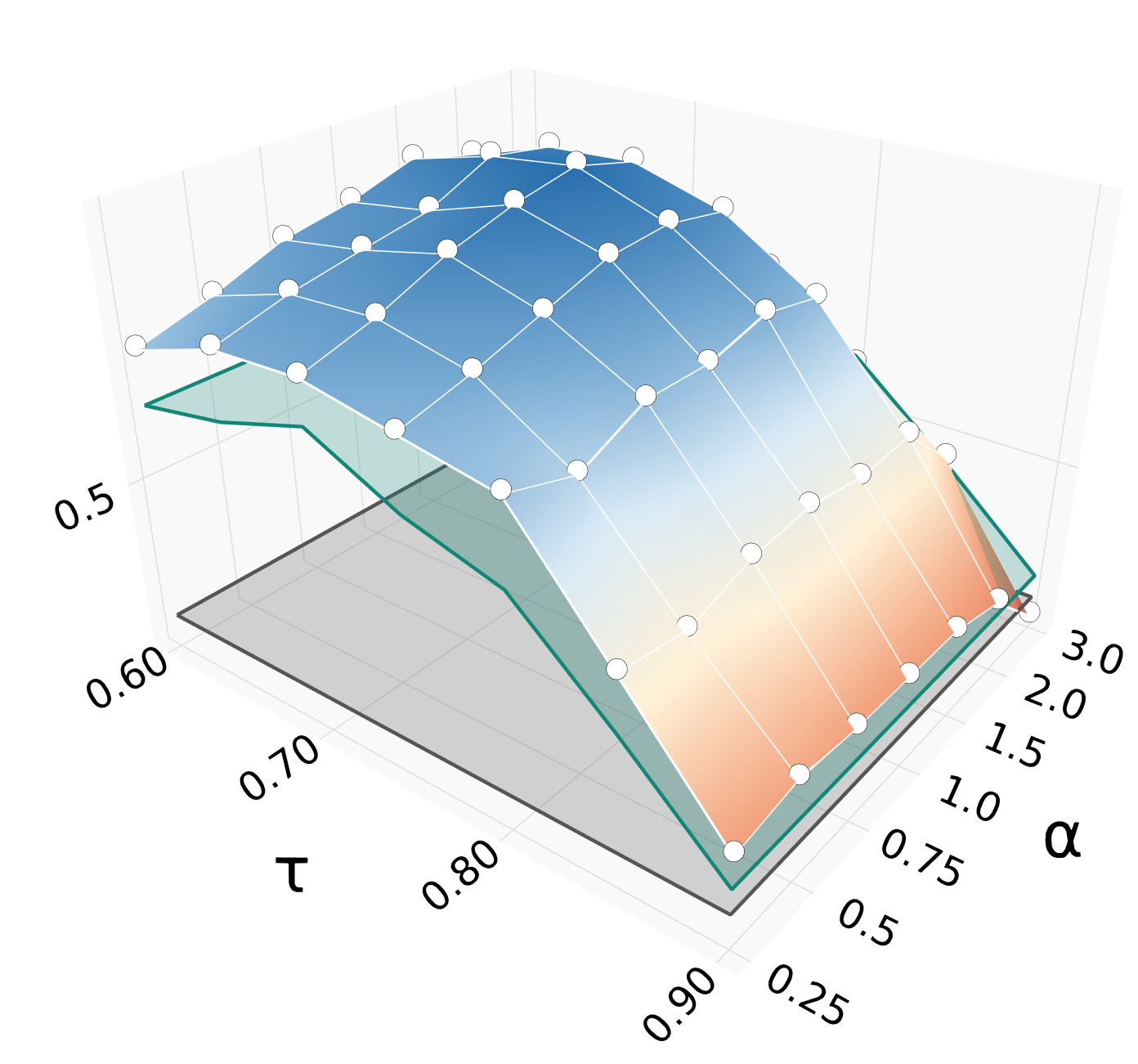}
\end{minipage}\hspace{0.05\linewidth}%
\begin{minipage}[t]{0.27\linewidth}\centering {\scriptsize F1@0.5}\\[1pt]
\includegraphics[width=\linewidth]{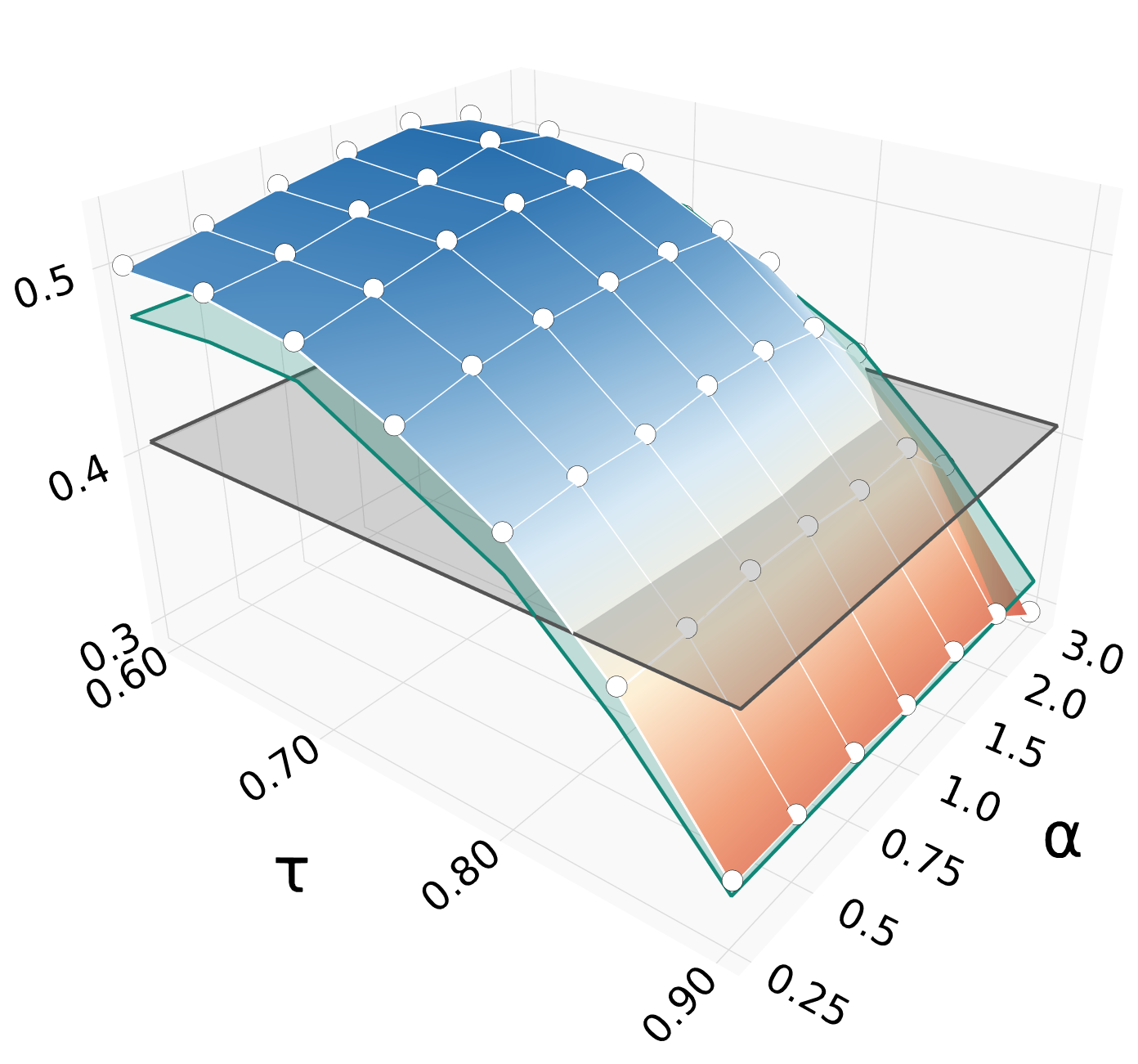}
\end{minipage}\hspace{0.05\linewidth}%
\begin{minipage}[t]{0.27\linewidth}\centering {\scriptsize GREEN}\\[1pt]
\includegraphics[width=\linewidth]{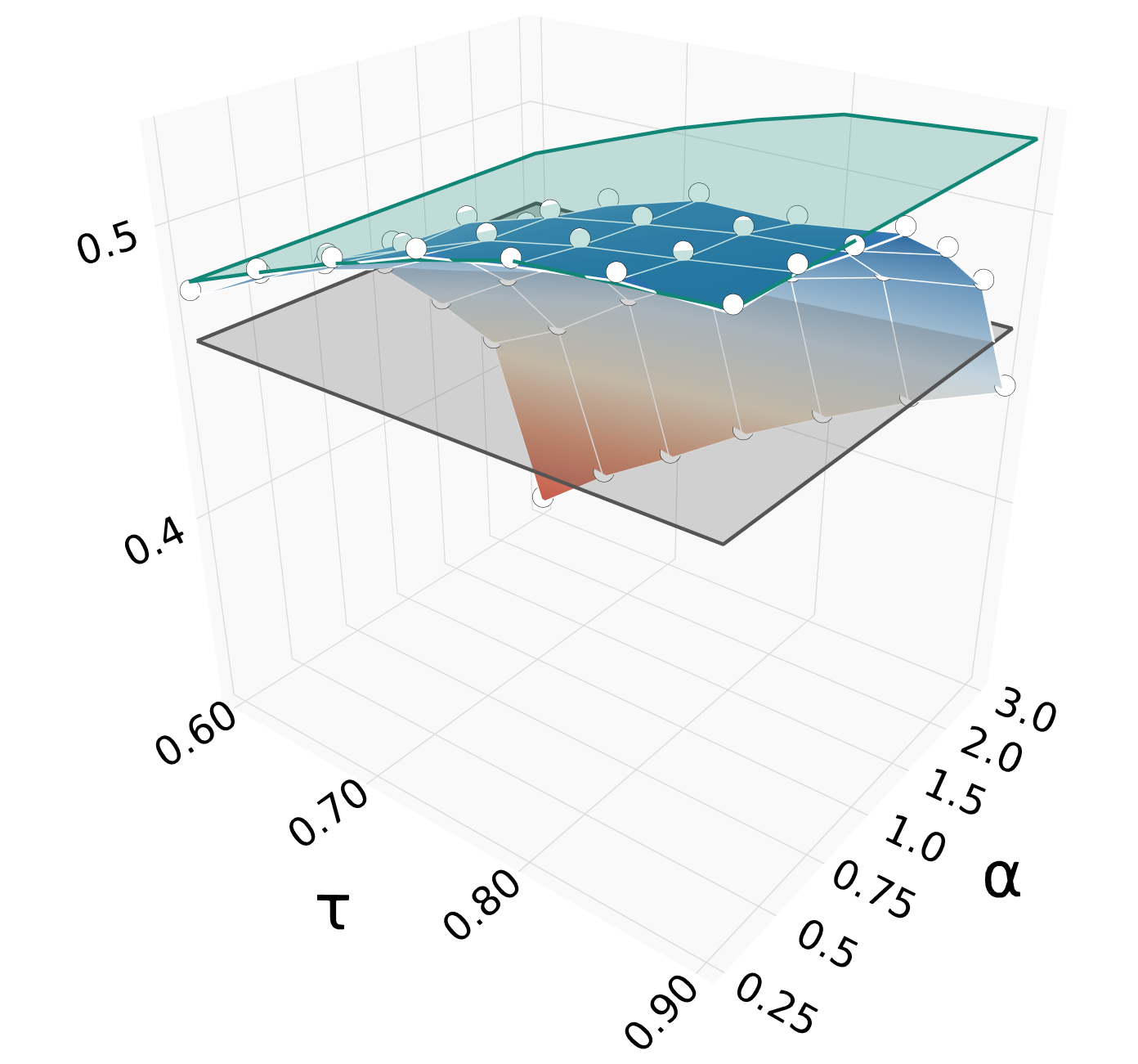}
\end{minipage}\\[2pt]
{\scriptsize COLIPRI with MedGemma-27B}\\[1pt]
\begin{minipage}[t]{0.27\linewidth}\centering {\scriptsize $\mathrm{F1_{max}}$}\\[1pt]
\includegraphics[width=\linewidth]{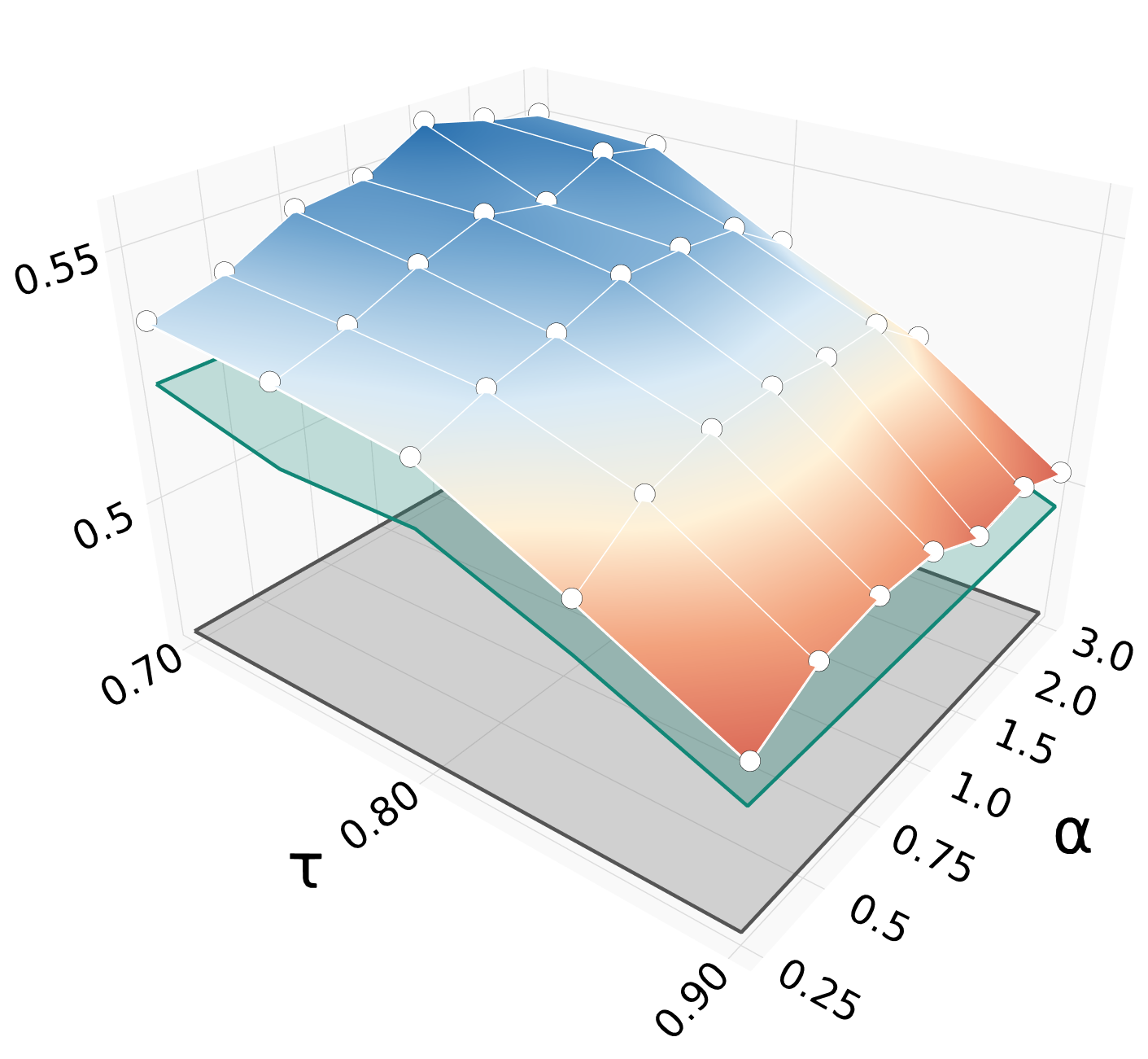}
\end{minipage}\hspace{0.05\linewidth}%
\begin{minipage}[t]{0.27\linewidth}\centering {\scriptsize F1@0.5}\\[1pt]
\includegraphics[width=\linewidth]{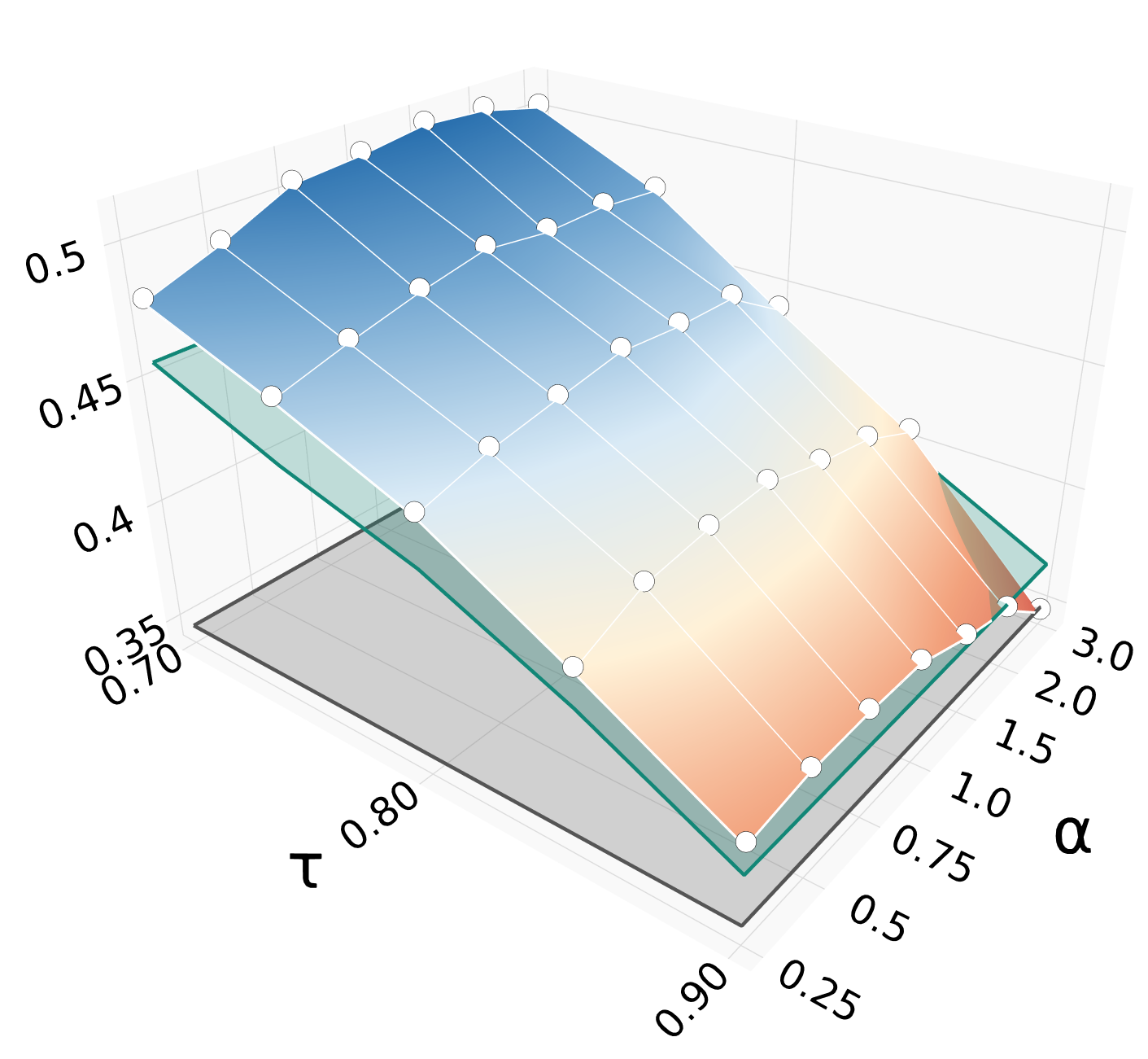}
\end{minipage}\hspace{0.05\linewidth}%
\begin{minipage}[t]{0.27\linewidth}\centering {\scriptsize GREEN}\\[1pt]
\includegraphics[width=\linewidth]{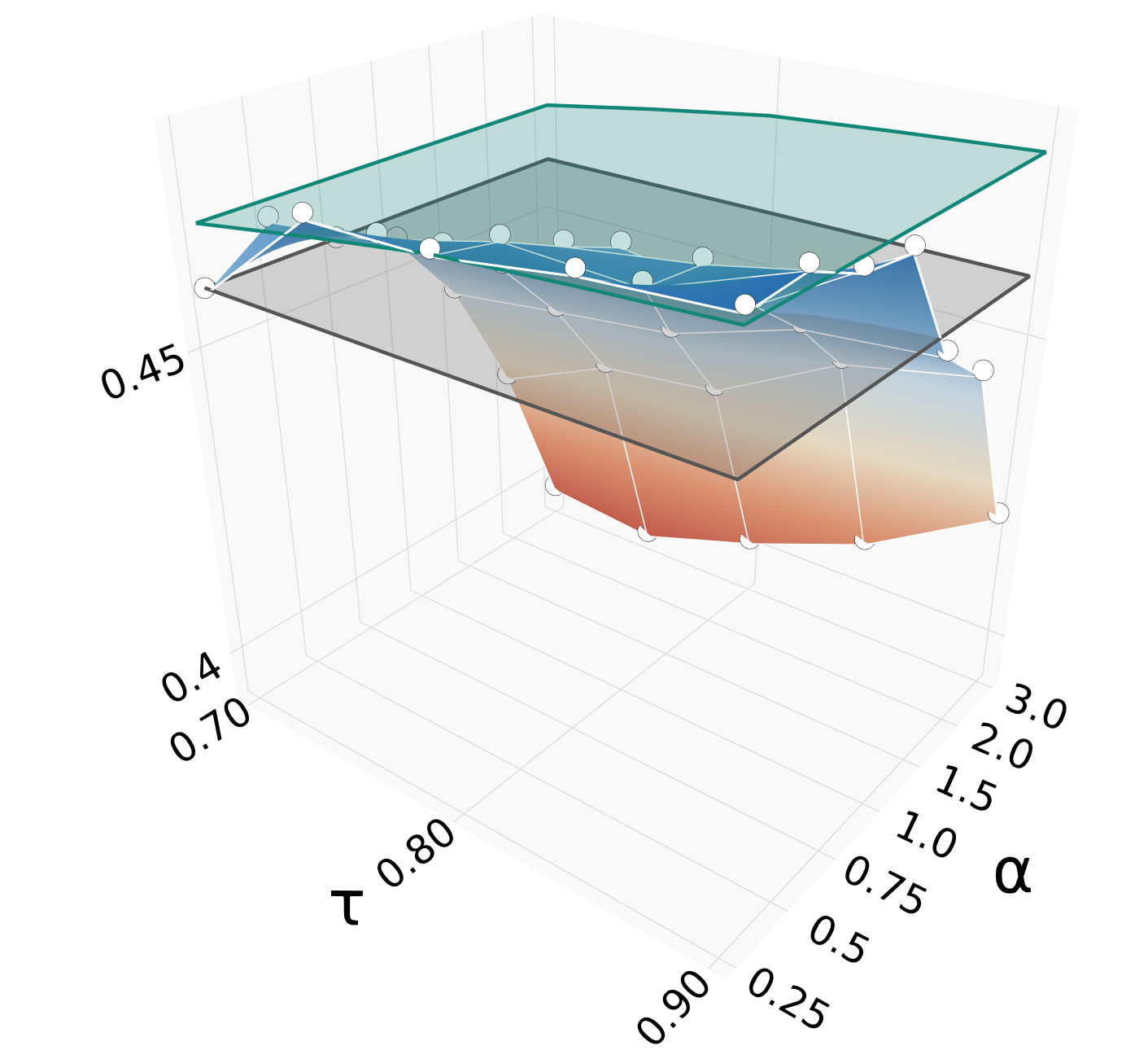}
\end{minipage}\\[2pt]
\includegraphics[width=0.36\linewidth]{figures/grid3d_legend.png}
\caption{SelfCue with the gate on empty conditions over $\alpha$ and $\tau$ on the other two pairs. The threshold is swept from $0.70$ on MedGemma-27B.}
\label{fig:grid3d_rest}
\end{figure}

Figure~\ref{fig:grid3d_rest} sweeps $\alpha$ and $\tau$ on the other two pairs. Every cell of both grids clears its own $M_{\mathrm{base}}$ and the same weights decoded at $\alpha = 0$ on $\mathrm{F1_{max}}$, and $\tau$ again moves the metric more than $\alpha$ does. The peak sits at $\tau = 0.70$ on VISD+CT-CLIP rather than the $0.75$ of the other two.

\subsection{The oracle ceiling}
\label{app:provenance}

The condition SelfCue decodes on comes from the probe, so the method inherits the probe's errors. Table~\ref{tab:oracle} reads it from the ground-truth labels instead, with the decoding rule, the checkpoint and the gate unchanged. Macro F1@0.5 rises from $0.480$ to $0.869$, so what limits the method is the readout rather than the rule that acts on it. GREEN stays where it was. A condition that is right about every abnormality buys nothing from the judge. This run was made on one pair.

\begin{table}[h]
\centering
\scriptsize
\begin{tabular}{lcccc}
\toprule
 & F1@0.5 & $\mathrm{F1_{max}}$ & AUROC & GREEN \\
\midrule
$M_{\mathrm{base}}$ & $0.395$ & $0.482$ & $0.739$ & $0.466$ \\
SelfCue, probe condition & $0.480$ & $0.542$ & $0.775$ & $0.510$ \\
SelfCue, labels as the condition & $0.869$ & $0.873$ & $0.937$ & $0.511$ \\
\bottomrule
\end{tabular}
\caption{\textbf{The oracle ceiling.} The same decoding rule with the condition read from the probe and from the ground-truth labels, on COLIPRI with Llama-3.1-8B.}
\label{tab:oracle}
\end{table}

\subsection{Student training choices}
\label{app:studentinit}

Two choices in training the student are ablated in Table~\ref{tab:student}, where it starts and which of the teacher's reports it sees. Where it starts moves F1@0.5 by less than $0.01$, and a student trained from scratch comes within $0.01$ of one warm-started from $M_{\mathrm{cond}}$. Which reports it sees matters more at the bottom than at the top. Keeping only the teacher's worst half costs $0.016$ against the whole corpus, while keeping only its best half gains nothing, and the two random halves place that comparison against corpus size rather than selection. Dropping the reports the teacher scores below $0.2$ is the one variant that improves on the whole corpus, by $0.003$ F1@0.5 and $0.005$ $\mathrm{F1_{max}}$. Every run here was trained on the earlier teacher corpus, written at $\alpha = 0.5$ with no threshold, so the rows compare with each other and not with the systems the paper reports.

\begin{table}[h]
\centering
\scriptsize
\begin{tabular}{lccc}
\toprule
 & F1@0.5 & $\mathrm{F1_{max}}$ & AUROC \\
\midrule
\multicolumn{4}{l}{\emph{Where the student starts}} \\
From $M_{\mathrm{cond}}$ & $0.483$ & $0.517$ & $0.770$ \\
From $M_{\mathrm{base}}$ & $0.476$ & $0.515$ & $0.772$ \\
From the stage-1 projector & $0.479$ & $0.517$ & $0.772$ \\
From scratch & $0.474$ & $0.510$ & $0.769$ \\
\midrule
\multicolumn{4}{l}{\emph{Which teacher reports it sees}} \\
The whole corpus & $0.483$ & $0.517$ & $0.770$ \\
The highest-scoring half & $0.481$ & $0.519$ & $0.775$ \\
The lowest-scoring half & $0.467$ & $0.510$ & $0.767$ \\
A random half & $0.478$ & $0.520$ & $0.774$ \\
Another random half & $0.476$ & $0.516$ & $0.774$ \\
Teacher F1 at least $0.2$ & $0.486$ & $0.522$ & $0.773$ \\
Teacher F1 at least $0.5$ & $0.477$ & $0.516$ & $0.772$ \\
\bottomrule
\end{tabular}
\caption{\textbf{Student variants.} All rows are on COLIPRI with Llama-3.1-8B, and the first row of each block is the choice the paper makes.}
\label{tab:student}
\end{table}


\end{document}